\documentclass[10pt,twocolumn,letterpaper]{article}

\usepackage{iccv}
\usepackage{times}
\usepackage{epsfig}
\usepackage{graphicx}
\usepackage{amsmath}
\usepackage{amssymb}

\usepackage{booktabs}
\usepackage{enumitem}
\usepackage{multirow}
\usepackage{soul}
\usepackage{pifont} 
\usepackage[T1]{fontenc}
\usepackage{mathabx}
\usepackage{makecell}
\usepackage[dvipsnames]{xcolor}
\usepackage{color, colortbl}
\usepackage{caption}

\newcommand{\blue}[1]{{\textbf{\color{blue}#1}}}
\newcommand\Tstrut{\rule{0pt}{2.3ex}}         

\def\CircleArrowright{\ensuremath{%
  \rotatebox[origin=c]{310}{$\circlearrowright$}}}

\newcommand{\vlnbert}{VLN$\protect\CircleArrowright$BERT}
\newcommand{\ours}{ScaleVLN}

\usepackage[pagebackref=true,breaklinks=true,letterpaper=true,colorlinks,bookmarks=false]{hyperref}

\iccvfinalcopy 

\def\iccvPaperID{4534} 

\ificcvfinal\fi

\begin{document}

\title{Scaling Data Generation in Vision-and-Language Navigation}

\author{Zun Wang$^{*\spadesuit1,2}$ \quad Jialu Li$^{*3}$ 
\quad Yicong Hong$^{*\dag1}$ \\
Yi Wang$^2$ \quad Qi Wu$^4$ \quad Mohit Bansal$^3$ \quad Stephen Gould$^1$ \quad
Hao Tan$^5$ \quad Yu Qiao$^2$ \\
$^1$The Australian National University\quad$^2$OpenGVLab, Shanghai AI Laboratory\quad\\
$^3$UNC, Chapel Hill\quad$^4$University of Adelaide\quad
$^5$Adobe Research\\ 
{\tt\small wangzun@pjlab.org.cn, jialuli@cs.unc.edu, mr.yiconghong@gmail.com} \\ 
{\tt\small  Project URL: \url{https://github.com/wz0919/ScaleVLN}}
}

\maketitle
\ificcvfinal\thispagestyle{empty}\fi

\begin{abstract}
Recent research in language-guided visual navigation has demonstrated a significant demand for the diversity of traversable environments and the quantity of supervision for training generalizable agents. To tackle the common data scarcity issue in existing vision-and-language navigation datasets, we propose an effective paradigm for generating large-scale data for learning, which applies 1200+ photo-realistic environments from HM3D and Gibson datasets and synthesizes 4.9 million instruction-trajectory pairs using fully-accessible resources on the web. Importantly, we investigate the influence of each component in this paradigm on the agent's performance and study how to adequately apply the augmented data to pre-train and fine-tune an agent. Thanks to our large-scale dataset, the performance of an existing agent can be pushed up (+11\% absolute with regard to previous SoTA) to a significantly new best of 80\% single-run success rate on the R2R test split by simple imitation learning. The long-lasting generalization gap between navigating in seen and unseen environments is also reduced to less than 1\% (versus 8\% in the previous best method). 
Moreover, our paradigm also facilitates different models to achieve new state-of-the-art navigation results on CVDN, REVERIE, and R2R in continuous environments.

{\let\thefootnote\relax\footnote{$^{*}$Equal contribution. $^{\dag}$Project lead.}}
{\let\thefootnote\relax\footnote{ $^{\spadesuit}$Research done during internship at Shanghai AI Lab.}}

\end{abstract}


\section{Introduction}
\label{sec:intro}

Vision-and-Language Navigation (VLN)~\cite{anderson2018r2r} is a challenging task that requires an agent to navigate in photo-realistic environments, following human natural language instructions such as ``\textit{Walk downstairs, move towards the dining table, turn left to the kitchen, and stop in front of the fridge.}'' Addressing VLN relies heavily on correctly interpreting the instructions, perceiving the environments, and learning from interaction, which demands a large amount of diverse visual-language data for learning. Recent research shows that scaling up the diversity of environments and the quantity of demonstration for training VLN agents are promising in improving generalization to unseen scenes~\cite{chen2022hm3dlearning,kamath2022marval}. Compared to previous approaches of addressing data scarcity by augmenting agent's observations~\cite{li2022envedit,tan2019envdrop} or employing large vision-linguistic models pre-trained with image-text data from the web~\cite{guhur2021airbert,hong2020recurrent,majumdar2020improving,shah2022lmnav,shen2021benefit}, utilizing additional traversable environments allows the agents to learn from in-domain visual-language data and physical interaction in the space.

\begin{figure}[t]
  \centering
    \includegraphics[width=1.0\columnwidth]{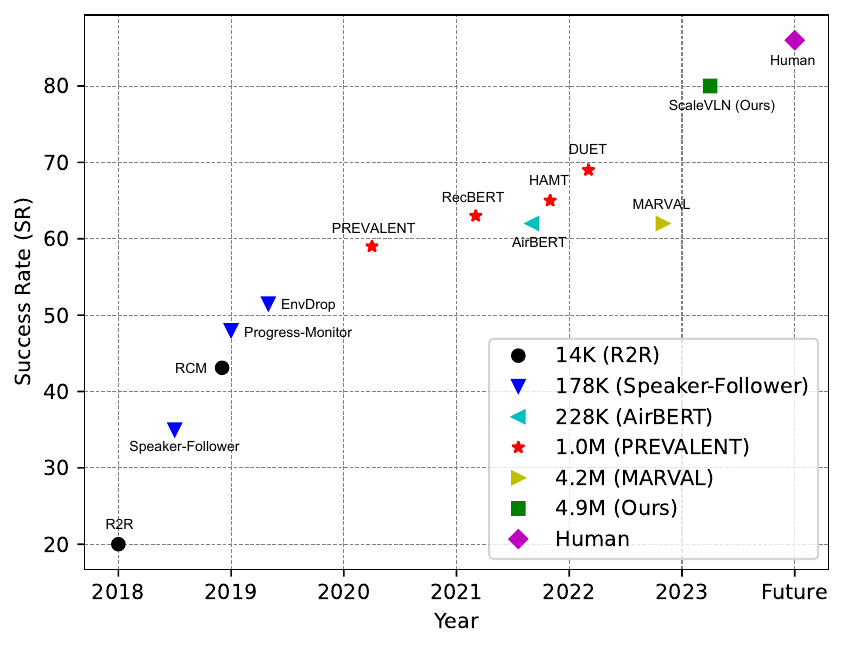}
    \vspace{-20pt}
  \caption{Agent success rate with increasing data size on addressing the R2R navigation task. Our proposed method creates 4.9M instruction-trajectories pairs for learning, which greatly boosts the agent's performance, and for the first time approaching human results.}
  \label{fig:vln_trend}
  \vspace{-15pt}
\end{figure}

In light of this, recent large datasets which contain hundreds of interactive scenes have been created~\cite{deitke2022procthor,ramakrishnan2021hm3d,xia2018gibson}, as well as a vast amount of human demonstrations have been collected~\cite{anderson2020rxr,ramrakhya2022habweb} for learning visual navigation, leading to significant improvement in agent's performance.
However, the process towards such large-scale training involves solving a series of key sub-problems such as how to build navigation graphs~\cite{anderson2018r2r,hong2022bridging,kamath2022marval}, how to recover corrupted rendered images~\cite{anderson2020sim,krantz2022sim}, and how to generate navigational instructions~\cite{dou2022foam,fried2018speaker,wang2022less,wu2021improvedspeaker}, which significantly influence the quality of collected data and should be investigated thoroughly. Meanwhile, an agent capable of understanding human natural language and navigating in photo-realistic environments is a complex and modularized system~\cite{an2022bevbert,chen2022sevol,chen2022duet,hong2023ego2map,wang2021structured,wang2019reinforced,zhou2023navgpt,zhu2021scoa}, and it is important to study how to effectively utilize the large-scale data to benefit the training of navigational agents adequately.

In this paper, we propose an effective paradigm for large-scale vision-and-language navigation (VLN) training and quantitatively evaluate the influence of each component in the pipeline. Specifically, we utilize environments in both the HM3D~\cite{ramakrishnan2021hm3d} and the Gibson~\cite{xia2018gibson} datasets, build navigation graphs for the environments based on the Habitat simulator~\cite{savva2019habitat}, sample new trajectories and generate corresponding instructions~\cite{tan2019envdrop}, and train agents~\cite{chen2021hamt,chen2022duet} for solving downstream navigation tasks~\cite{anderson2018r2r,jain2019stay,krantz2020navgraph,qi2020reverie,thomason2020cvdn}. Different from previous methods such as AutoVLN~\cite{chen2022hm3dlearning} and MARVAL~\cite{kamath2022marval}, we build navigation graphs using an excessive viewpoint sampling and aggregation algorithm, following the graph construction heuristic proposed in~\cite{hong2022bridging}, which results in fully-connected graphs with high coverage in open space. Additionally, we address the issue of corrupted rendered images from HM3D and Gibson environments with the Co-Modulated GAN~\cite{zhao2021comodgan}, which we train to generate photo-realistic images from the faulty rendered images with broken, distorted, or missing regions, to mitigate the noise in visual data.
Unlike MARVAL, which uses a non-public language generation model Marky~\cite{wang2022less} and visual encoder MURAL~\cite{jain2021mural}, as well as synthesizes observations from novel viewpoints with an image-to-image GAN~\cite{koh2022simple}, our large-scale training regime is fully reproducible and straightforward to execute, while leading to a significant improvement on agent's performance.

Through comprehensive experiments, we find that a fully traversable navigation graph is crucial to improve the agent's performance for downstream tasks with detailed instructions like R2R. Besides, we show that recovering photo-realistic images from the rendered images is very beneficial, especially for the low-quality 3D scans from the Gibson environments. Results also suggest that an agent can consistently benefit from having more diverse visual data, and learning from additional scenes helps agents to generalize better to unseen environments than simply learning from more data. Moreover, we validate that an agent trained with augmented instructions generated by a simple LSTM-based  model~\cite{tan2019envdrop} can achieve good performance on multiple navigation tasks ~\cite{anderson2018r2r,qi2020reverie,thomason2020cvdn}. Last but not least, we find that appropriately combining our augmented data with the original data in pre-training and fine-tuning can benefit the agent's generalization ability. 

Remarkably, by following the above analysis as data augmentation and agent training guidelines, our resulting VLN model achieves 80\% success rate (SR) on the R2R test split by simple imitation learning without pre-exploration~\cite{fu2020counterfactual,tan2019envdrop,zhu2020vision}, beam search~\cite{fried2018speaker,majumdar2020improving,xie2022beam} or model ensembling~\cite{qin2021ensemble}, and successfully eliminates the gap between navigating in seen and unseen environments. This result significantly outperforms previous best method (73\%)~\cite{an2022bevbert}, 
and reduces the difference towards human performance (86\% SR\footnote{Note that human followers only have egocentric views, while our model follows the common approach of applying panoramic observations.})~\cite{anderson2018r2r} to 6\%. Our method also achieves new state-of-the-art results on different language-guided visual navigation problems, including CVDN~\cite{thomason2020cvdn} and REVERIE~\cite{qi2020reverie}. 
Moreover, although the augmented data is discrete, it helps boost VLN performance in continuous environments (R2R-CE)~\cite{an20221st,hong2022bridging,krantz2020navgraph}, a much more realistic but difficult scenario, by 5\% SR. All the results demonstrate the great effectiveness and generalization potential of our training regime. 
In summary, our main contributions include:
\begin{enumerate}
[itemsep=0.1em,parsep=0em,topsep=0em,partopsep=0em,leftmargin=2em]
    \item A simple, effective, fully automated and reproducible large-scale training paradigm for vision-and-language navigation.
    \item Comprehensive analysis of the entire data augmentation pipeline and utilizing the large data for training.
    \item New state-of-the-art results on navigation tasks including R2R, CVDN, REVERIE, and R2R-CE.
\end{enumerate}


\section{Related Works}
\label{sec:rel_works}

\paragraph{Vision-and-Language Navigation}
Learning to navigate in unvisited environments following natural language instructions is an important step toward intelligent robots that can assist humans with daily activities. In the past years, a great variety of scenarios have been proposed for VLN research, such as navigation with comprehensive language guidance~\cite{anderson2018r2r,jain2019stay,anderson2020rxr}, navigation by interpreting dialog history~\cite{de2018talk,padmakumar2022teach,thomason2020cvdn}, grounding remote objects with high-level instructions~\cite{qi2020reverie,zhu2021soon}, and navigation in continuous environments that closely approximate the real world~\cite{krantz2022iterative,krantz2020navgraph}. To address the problem, early research mainly focuses on developing task-specific models and training methods to better exploit visual-textual correspondence for decision making~\cite{an2021neighbor,anderson2019chasing,deng2020evolving,ke2019tactical,lin2022multimodal,ma2019self,qi2021road,qi2020object,wang2020active}.

\paragraph{Large-Scale Visual Navigation Learning}
Due to the expensive navigational data collection process, learning to navigate usually faces a data scarcity issue~\cite{habitat2020sim2real,anderson2018r2r,batra2020rearrangement, batra2020objectnav,ehsani2021manipulathor,anderson2020rxr,qi2020reverie,thomason2020cvdn}. Many works have been proposed to scale up the training data by collecting more human annotations~\cite{ramrakhya2022habweb} or creating new environments~\cite{deitke2022procthor,ramakrishnan2021hm3d}. Moreover, recent studies tend to establish a scalable regime, utilizing extensive automatically-generated data to push the limit of agent performance~\cite{chen2022hm3dlearning,kamath2022marval}, or introducing large-scale pre-training approaches to improve the generalizing ability~\cite{chen2021hamt,li2023vlnsig,qiao2023hop+}. In this paper, we create a simple paradigm for scaling VLN training, and through comprehensive analysis, we seek a valuable guideline for data acquisition and agent training for future research.

\section{Scaling Data for Learning VLN}
\label{sec:methods}

We outline the necessary resources for learning VLN, followed by the details of our method for creating the large-scale augmented dataset from additional environments. Note that in this section, we only present our method to generate instruction-path pairs in R2R-style, which will be shared to address downstream R2R~\cite{anderson2018r2r}, CVDN~\cite{thomason2020cvdn} and R2R-CE~\cite{krantz2020navgraph} tasks. We refer to the \textit{Appendix}~\ref{sec_1} for data collection and model training details for REVERIE, whose data requires trajectories that lead to a specific object~\cite{qi2020reverie}.

\subsection{Resources for VLN Training}  

Most existing research on VLN is established over the discrete Matterport 3D environments (MP3D)~\cite{chang2017matterport3d} where an agent's positions and observations are constrained on viewpoints of predefined navigation graphs. The trajectory-instruction pairs are sampled and annotated based on these discrete graphs. Compared to navigation in continuous environments~\cite{krantz2020navgraph,savva2019habitat}, such simplification enables efficient learning and execution while remaining to be practical, because, essentially, VLN agents make decisions by executing a vision-and-language grounding process~\cite{anderson2018r2r}. There are also some recent works that attempt to transfer agents designed for discrete scenarios to continuous environments~\cite{anderson2021sim,hong2022bridging,krantz2021waypoint,krantz2022sim}. Our data augmentation paradigm produces discrete supervisions, whereas we show in experiments that it also facilitates VLN learning in continuous scenes. In summary, scaling VLN data typically requires collecting new visual environments, discretizing the environments by building navigation graphs, sampling trajectories (sequences of images) on the graphs, and generating corresponding instructions. Following this procedure, we specify our data augmentation paradigm below.

\begin{figure*}[t]
  \centering
    \includegraphics[width=\textwidth]{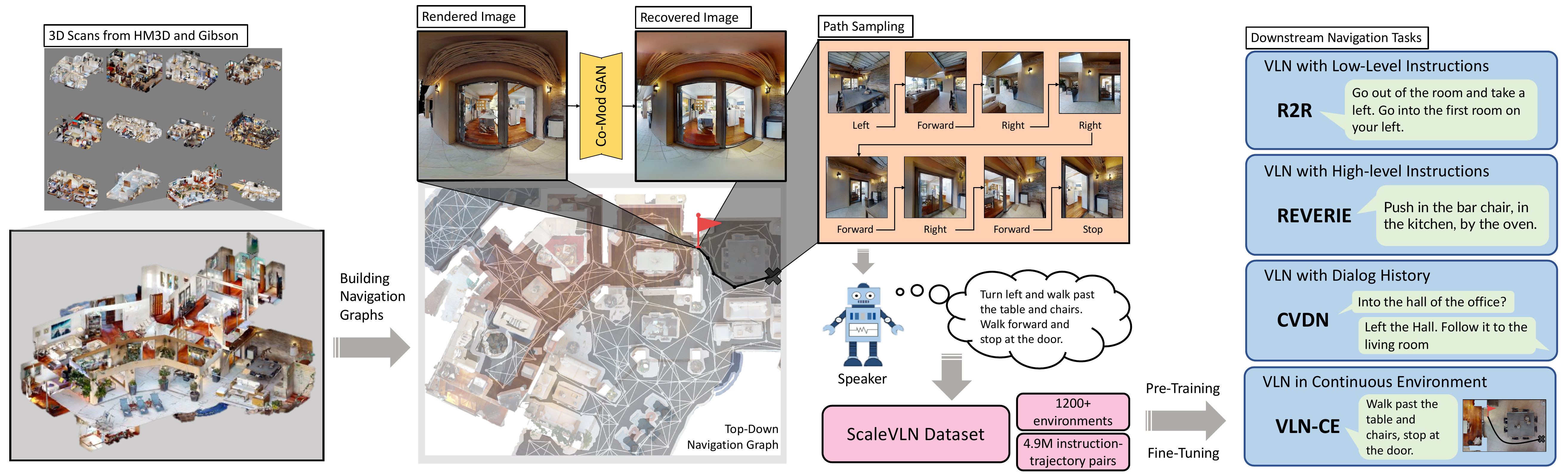}
  \caption{Our proposed paradigm (\ours) for generating large-scale augmented VLN data. \ours{} applies 1200+ unannotated 3D scans from the HM3D~\cite{ramakrishnan2021hm3d} and Gibson~\cite{xia2018gibson} environments, builds navigation graphs for each scene, recovers faulty rendered images with a Co-Mod GAN~\cite{zhao2021comodgan}, samples trajectories and generates corresponding instructions, resulting in 4.9M augmented data to facilitate learning various downstream language-guided navigation tasks~\cite{anderson2018r2r,krantz2020navgraph,qi2020reverie,thomason2020cvdn}.}
  \label{fig:overall_pipeline}
\end{figure*}

\subsection{Generating Augmented Data}

\paragraph{Collecting Environments}
We adopt environments from HM3D~\cite{ramakrishnan2021hm3d} and Gibson~\cite{xia2018gibson} as the source of our visual data. Both datasets contain abundant, traversable, and simulated indoor 3D scans collected from real-world buildings, which support the learning of various visual navigation problems~\cite{batra2020objectnav,krantz2020navgraph,mezghani2021imagenav,savva2019habitat}. Specifically, we employ 800 training scenes from HM3D, and 491 training and validation scenes from Gibson (same as MARVAL~\cite{kamath2022marval}), resulting in more than 150k $m^2$ navigable area, which is around $\times$7.5 times larger than the training scenes of downstream MP3D environments (20k $m^2$, 61 scans).

\paragraph{Constructing Navigation Graphs}
We argue that a high-quality navigation graph needs to satisfy a number of criteria, including high coverage of the space to maximize visual diversity and fully traversable edges in appropriate lengths with nodes positioned close to the center of open space for sampling reasonable trajectories. Previous work AutoVLN~\cite{chen2022hm3dlearning} builds graphs with very sparse nodes and with edges that go across obstacles, limiting the quantity of sampled data and leading to impractical trajectories, while MARVAL~\cite{kamath2022marval} trains a model to predict navigable directions, which could make errors and overcomplicate the problem. In this work, we propose a very simple but accurate heuristic for building the graph: we first apply the existing navigable position sampling function in Habitat simulator~\cite{savva2019habitat} to sample an excessive amount of viewpoints which almost covers the entire open space while limiting the geodesic distance between any two viewpoints to be greater than 0.4 $m$. Then, we apply the Agglomerative Clustering algorithm to group adjacent viewpoints to a single viewpoint with a distance threshold of 1.0 $m$, automatically producing positions close to the center of open space. We create a rough graph by randomly connecting viewpoints within 5.0 $m$ separation while capping the maximal edges of a viewpoint to be five, and use the existing graph refinement approach~\cite{hong2022bridging} to obtain the final fully-connected and fully-traversable navigation graphs. We use this method to construct graphs for the 800+491 environments; the average edge length in the graphs is 1.41 $m$, and the average node degree is 4.55. We visualize the graphs in \textit{Appendix}~\ref{sec_3}.

\paragraph{Recovering Faulty Rendered Images} 
Although HM3D and Gibson provide a large amount of diverse indoor environments, the quality of the images rendered from their 3D meshes is often much worse than the camera-captured images (as shown in Figure~\ref{fig:overall_pipeline}). Previous work has shown that navigation agents trained with rendered views will perform significantly worse than agents trained with high-quality images~\cite{krantz2022sim}. As a result, we consider recovery of faulty rendered images as a process in our \ours{} paradigm.

We formulate this task as an image-to-image translation problem, where the model takes a rendered image as input and learns to recover the broken, distorted, or missing regions. Specifically, we adopt the Co-Modulated GAN (Co-Mod GAN)~\cite{zhao2021large}, a generative model that can leverage conditional information and retain the stochastic in unconditional generation. 
We train Co-Mod GAN on the rendered-and-camera-image pairs in Matterport3D datasets and use the trained model to recover the rendered images in HM3D and Gibson environments.

\paragraph{Sample Trajectories}

We sample trajectories on the navigation graphs of HM3D and Gibson environments. For navigation tasks with detailed instructions, we follow PREVALENT~\cite{hao2020prevalent} and collect all possible shortest routes between any two viewpoints connected by three to five intermediate nodes. This sampling strategy yields a total of 2,890,267 paths and 2,051,443 paths for the HM3D and Gibson environments, respectively.

\paragraph{Generate Navigational Instructions}

Finally, we apply the off-the-shelf EnvDrop Speaker~\cite{tan2019envdrop}, a simple LSTM-based language generation model trained on instruction-path pairs in R2R~\cite{anderson2018r2r}, to produce a navigational instruction for each sampled trajectory for navigation tasks with detailed instructions. Compared to the more powerful language model GPT-2~\cite{radford2019gpt2}, EnvDrop Speaker generates less diverse descriptions, but the resulting data can lead to similar improvement on agents addressing R2R task (see \S\ref{subsec:scale_data}).

\vspace{5pt}

Following the procedure above, our large-scale data augmentation paradigm creates 4,941,710 instruction-trajectory pairs for learning VLN. This size is $\times$352 larger than the R2R dataset and $\times$4.62 larger than the commonly applied augmented PREVALENT dataset~\cite{hao2020prevalent}.
\section{Experiments}
\label{sec:exps}

In this section, we present a comprehensive evaluation of the effect of each component in our data augmentation paradigm, investigate how to appropriately use the data for learning, and test agents~\cite{chen2022duet,chen2021hamt} pre-trained with our data on multiple VLN downstream tasks~\cite{anderson2018r2r,krantz2020navgraph,qi2020reverie,thomason2020cvdn}.

\subsection{Experimental Setup}

\paragraph{Datasets} 
We perform analysis mainly on the R2R dataset~\cite{anderson2018r2r}, while evaluating the generalization potential of our augmented data on REVERIE~\cite{qi2020reverie}, CVDN~\cite{thomason2020cvdn} and R2R-CE~\cite{krantz2020navgraph}. The datasets are outlined as follows:

\begin{itemize}[itemsep=0.1em,parsep=0em,topsep=0em,partopsep=0em,leftmargin=2em]
    \item \textbf{R2R} consists of 22k human-annotated navigational instructions, each describing a trajectory that traverses multiple rooms in MP3D~\cite{chang2017matterport3d}. 
    On average, an instruction contains 32 words, and each ground-truth path is formed by seven nodes with a total length of 10 $m$.
    \item \textbf{REVERIE} inherits the trajectories in R2R but provides high-level instructions which describe a target object. The task for an agent is first to find the object, and localize it in observation.
    \item \textbf{CVDN} provides dialogues between a navigator who tries to find a target by asking for guidance and an oracle with a privileged view of the best next step. An agent who addresses the task needs to find the way by interpreting the dialogue history.
    \item \textbf{R2R-CE} transfers the discrete trajectories in R2R to continuous 3D scans rendered by Habitat simulator~\cite{savva2019habitat}, where an agent can freely travel in the open space and need to interact with obstacles. The dataset contains 16k instruction-trajectory pairs after removing non-transferable paths.
\end{itemize}

Besides, we also adopt the widely applied augmented R2R dataset PREVALENT~\cite{hao2020prevalent} in our experiments, which only has 178,270 samples created from MP3D scenes. For simplicity, we use PREV, HM-E, and Gib-E to denote PREVALENT data, our augmented data from HM3D and Gibson scenes with instructions generated by EnvDrop Speaker~\cite{tan2019envdrop}, respectively, and use the term \ours{} data for all our HM-E and Gib-E data in the following sections.

\paragraph{Baseline VLN Models}

We employ the recently proposed VLN agents, Dual-Scale Graph Transformer (DUET)~\cite{chen2022duet} and History Aware Multimodal Transformer (HAMT)~\cite{chen2021hamt} as the baseline models in our experiments. The primary idea of DUET is to build a topological map on the fly, which extends the agent's action space from its current viewpoint to all navigable directions encountered during navigation, therefore, greatly facilitating planning and error correction. HAMT explicitly stores the observations at each navigational step, which benefits the learning of sequence-to-sequence alignment between vision and instruction. We refer readers to their original papers for more technical details. In our experiments, we apply the DUET agent in R2R, CVDN and REVERIE, whereas using the HAMT agent in R2R-CE, since the two models report the best results on these datasets, respectively.

\paragraph{Training} 

We use the augmented data for a two-stage VLN agent training, \ie, pre-training and fine-tuning. In pre-training, we consider the most widely applied proxy tasks in previous work, Masked Language Modeling (MLM), Masked Region Modeling (MRM), and Single-Action Prediction (SAP)~\cite{chen2021hamt,chen2022duet,hao2020prevalent,qiao2022hop,qiao2023hop+}, to enhance agent's language understanding, visual perception, and to benefit cross-modal grounding between instruction and observation (whose effect will be studied in \S\ref{subsec:utilize_data}). We refer to \textit{Appendix}~\ref{sec_1} for their implementation details.

After pre-training, similar to AutoVLN~\cite{chen2022duet} and MARVAL~\cite{kamath2022marval}, we fine-tune the model simply with imitation learning (IL) method DA{\footnotesize GGER}~\cite{ross2011dagger}. Specifically, at each time step, an agent performs an action sampled from the predicted probability of its action space, and minimizes the loss between the sampled action and the ground truth. This method allows an agent to learn from paths that cover wide space and reduces the exposure bias caused by teacher forcing~\cite{lamb2016professor}.

\paragraph{Implementation Details}

We apply CLIP ViT-B/16~\cite{radford2021clip}, a visual transformer~\cite{dosovitskiy2020vit} pre-trained to align millions of image-text pairs from the web, as the visual encoder in all our experiments if not specified otherwise. We refer to \textit{Appendix}~\ref{sec_1} for more details.

To address R2R and CVDN, we pre-train DUET for 20k iterations with a batch size of 256 and learning rate of $5 \times 10^{-5}$ on two NVIDIA Tesla A100 GPUs for about 72 GPU hours. 
We select one of the models logged in pre-training for fine-tuning according to the accuracy in solving proxy tasks and the performance in following R2R instructions. The selected model is then fine-tuned for 200k iterations with batch size 16 on a single GPU on both R2R and ScaleVLN datasets, which takes about 48 GPU hours to reach the peak performance.
For CVDN, we directly fine-tune the pre-trained DUET model on the dataset with the same configurations as fine-tuning on R2R.
For R2R-CE, we pre-train a single HAMT model with the same data and configurations, then fine-tune the model on the dataset. Similar to prior work~\cite{an2023etpnav,an20221st,wang2022internvideo}, our HAMT agent in R2R-CE leverages a candidate waypoint predictor~\cite{hong2022bridging} which predicts navigable locations to support agent's high-level decision-making process.

\paragraph{Evaluation Metrics}
 Standard metrics~\cite{anderson2018r2r} are applied to assess the agent's performance, including Trajectory Length (TL) which is the average length of the agent's predicted path in meters, Navigation Error (NE) which is the average distance between the agent's final position and the target in meters, Success Rate (SR) which is the ratio of agents that stop within 3 meters to the target viewpoint, and Success penalized by Path Length (SPL)~\cite{anderson2018spl}. For R2R-CE, normalized Dynamic Time Warping (nDTW) is an additional metric that measures the step-wise alignment between the ground truths and the agent-predicted paths~\cite{ilharco2019ndtw}. For CVDN, Goal Progress (GP) is the only metric; it measures the average difference between the length of the completed trajectory and the remaining distance to goal~\cite{thomason2020cvdn}.

\begin{table}[t]
  \begin{center}
  \resizebox{\columnwidth}{!}{
  \begin{tabular}{l|cc|ccc|rrr}
    \hline \hline
     \multicolumn{1}{c|}{\multirow{2}{*}{Methods}} & \multicolumn{2}{c|}{HM3D Nav Graphs} & \multicolumn{3}{c|}{R2R Val-Seen} &\multicolumn{3}{c}{R2R Val-Unseen} \\
     \cline{2-9} & \multicolumn{1}{c}{Density } & \multicolumn{1}{c|}{Collision} & \multicolumn{1}{c}{NE$\downarrow$} & \multicolumn{1}{c}{SR$\uparrow$} &
    \multicolumn{1}{c|}{SPL$\uparrow$} & \multicolumn{1}{c}{NE$\downarrow$} & \multicolumn{1}{c}{SR$\uparrow$} &
    \multicolumn{1}{c}{SPL$\uparrow$} \Tstrut\\
    \hline \hline
        None & --  & -- & 2.51 & 76.89 & 69.71 & 3.06 & 72.92 & 62.82\\
        \hline
     AutoVLN & 0.36 & 29.35\% & \textbf{1.90} & \textbf{84.43} & \textbf{79.10} & 3.08 & 72.75 & 62.56 \\
     
          Ours & 1.16 & 0.00\% & 2.25 & 79.82 & 75.06 & \textbf{2.75} & \textbf{76.01} & \textbf{66.94} \\
    \hline \hline
  \end{tabular}}
\end{center}
\vspace{-10pt}
\caption{Comparison on navigation graphs. \textit{Density} is computed as the number of nodes per navigable area (node/$m^2$), and \textit{Collision} is the ratio of edges that go through obstacles. \textit{None} means the agent only learn from R2R and PREV data). }
\label{tab:compare_graph}
\vspace{-10pt}
\end{table}

\subsection{Scale VLN Data, What Really Matters?}
\label{subsec:scale_data}

\paragraph{Effect of Navigation Graphs}
We first study the effect of different navigation graphs in  Table~\ref{tab:compare_graph}, where we compare the graphs from our method to AutoVLN~\cite{chen2022hm3dlearning}. For fairness, both methods only use 800 HM3D scenes without recovering faulty rendered images. We can see that generating augmented data from AutoVLN's graph cannot benefit the agent's performance in unseen environments. We suspect this is mainly due to a high ratio of edges that go through obstacles, resulting in noisy and misleading trajectories that do not exist in the downstream navigation graph.
On the contrary, our fully traversable graphs with a high density of viewpoints produce effective data, which greatly improves the results, suggesting the importance of graph quality in sampling discrete augmented data.

\begin{table}[t]
  \begin{center}
  \resizebox{\columnwidth}{!}{
  \begin{tabular}{rr|rrrr|rrrr}
    \hline \hline
     \multicolumn{2}{c|}{\makecell[c]{HM-E Aug}} & \multicolumn{4}{c|}{R2R Val-Seen} &\multicolumn{4}{c}{R2R Val-Unseen} \\
     \hline
     \#Scenes & \#Samples & \multicolumn{1}{c}{TL} & \multicolumn{1}{c}{NE$\downarrow$} & \multicolumn{1}{c}{SR$\uparrow$} &
    \multicolumn{1}{c|}{SPL$\uparrow$} & \multicolumn{1}{c}{TL} & \multicolumn{1}{c}{NE$\downarrow$} & \multicolumn{1}{c}{SR$\uparrow$} &
    \multicolumn{1}{c}{SPL$\uparrow$} \Tstrut\\
    \hline \hline
    800 & 2890k &   
    12.63 & 2.27 & 79.24 & 73.34 &
    12.83 & \textbf{2.62} & \textbf{76.59} & \textbf{67.74}\\
    800 & 1400k & 12.21 & 2.18 & 80.71 & 75.92 & 12.97 & 2.71 & 76.01 & 66.56 \\
    800 &  700k & 12.63 & \textbf{1.86} & \textbf{83.35} & \textbf{77.97} & 13.83 & 2.69 & 76.25 & 66.00\\
    400 & 700k & 
    12.76 & 1.87 & 82.96 & 77.05 &
     13.80 & 2.78 & 75.22 & 65.32
     \\
    200 &  700k &       
     11.95 & 1.95 & 82.66 & \textbf{77.97} &
    13.29 & 2.73 & 74.84 & 64.59 \\
    \hline
    0   &    0 &      13.28 & 2.51 & 76.89 & 69.71 &
     13.53 & 3.06 & 72.92 & 62.82\\
    \hline \hline
  \end{tabular}}
\end{center}
\vspace{-10pt}
\caption{Comparison of the quantity of augmented scene and samples. Here each experiment is pre-trained on data from R2R, PREV, and HM-E, and fine-tuned on R2R.}
\label{tab:quan_env_sample}
\end{table}

\paragraph{Effect of More Data }

Table~\ref{tab:quan_env_sample} shows the influence of the quantity of additional environments and training data. We can see that with the same amount of augmented scenes (800), agent performance in val-unseen gradually increases with higher sampling density.
On the other hand, generating the same amount of samples (700k) from more environments leads to better results. And it is clear that \#Scenes has a stronger impact than \#Samples, which suggests the importance of having more diverse environments for learning VLN.
Then, in Table~\ref{tab:add_gibson}, we further increase the quantity of training samples from Gibson environments for comparison and evaluate the pre-trained model's performance on R2R without fine-tuning. Conclusions from the previous table still hold: adding more scenes and data can bring a steady performance gain to the agent.

\begin{table}[t]
  \begin{center}
  \resizebox{\columnwidth}{!}{
  \begin{tabular}{l|c|rrr|rrr}
    \hline \hline
     \multicolumn{1}{c|}{\multirow{2}{*}{Pre-Train}} & \multicolumn{1}{c|}{\multirow{2}{*}{\makecell{Fine-Tune}}} & \multicolumn{3}{c|}{R2R Val-Seen} &\multicolumn{3}{c}{R2R Val-Unseen} \\
     \cline{3-8}  & & \multicolumn{1}{c}{NE$\downarrow$} & \multicolumn{1}{c}{SR$\uparrow$} &
    \multicolumn{1}{c|}{SPL$\uparrow$} & \multicolumn{1}{c}{NE$\downarrow$} & \multicolumn{1}{c}{SR$\uparrow$} &
    \multicolumn{1}{c}{SPL$\uparrow$} \Tstrut\\
    \hline \hline
    R2R, PREV & R2R & 2.51 & 76.89 & 69.71 & 3.06 & 72.92 & 62.82 \\    
    R2R, PREV + HM-E & R2R &  2.27 & 79.24 & 73.34 & 2.62 & 76.59 & 67.74 \\
    R2R, PREV + \ours{} & R2R & \textbf{2.02} & \textbf{80.51} & \textbf{74.88} & \textbf{2.53} & \textbf{78.08} & \textbf{68.31} \\
    \hline
    R2R, PREV & -- & 3.77 & 67.19 & 64.49 & 5.80 & 47.42 & 45.30 \\
    R2R, PREV + HM-E & -- & 4.04 & 64.64 & 62.00 & 5.03 & 55.09 & 52.23 \\
    R2R, PREV + \ours{} & -- & \textbf{3.64} & \textbf{71.11} & \textbf{68.53} & \textbf{4.90} & \textbf{57.00} & \textbf{54.03} \\
    \hline \hline
  \end{tabular}}
\end{center}
\vspace{-10pt}
\caption{Results of adding more augmented data and the pre-trained model performance without fine-tuning.}
\label{tab:add_gibson}
\vspace{-10pt}
\end{table}

\paragraph{Effect of Image Quality}

We evaluate the influence of image quality in augmented data on agent performance in Table~\ref{tab:recover_img}. We can see that for data from both HM3D and Gibson, recovering the rendered raw images can provide a noticeable benefit to the agent's results. Such improvement is more apparent for Gib-E because a large portion of 3D meshes for reconstructing Gibson scenes is in low-quality~\cite{xia2018gibson}, which leads to largely broken or distorted rendered views. In fact, HM-E (R) + Gib-E (F) leads to worse results than HM-E (R) alone even with 61\% more different environments, suggesting the great importance of having data with high visual quality.

\begin{table}[t]
  \begin{center}
  \resizebox{\columnwidth}{!}{
  \begin{tabular}{l|rrrr|rrrr}
    \hline \hline
     \multicolumn{1}{c|}{\multirow{2}{*}{Scenes}}  & \multicolumn{4}{c|}{R2R Val-Seen} &\multicolumn{4}{c}{R2R Val-Unseen} \\
     \cline{2-9} &  \multicolumn{1}{c}{TL} & \multicolumn{1}{c}{NE$\downarrow$} & \multicolumn{1}{c}{SR$\uparrow$} &
    \multicolumn{1}{c|}{SPL$\uparrow$} & \multicolumn{1}{c}{TL} & \multicolumn{1}{c}{NE$\downarrow$} & \multicolumn{1}{c}{SR$\uparrow$} &
    \multicolumn{1}{c}{SPL$\uparrow$} \Tstrut\\
    \hline \hline
    HM-E (F) & 12.17 & \textbf{2.25} & \textbf{79.82} & \textbf{75.06} & 12.64 & 2.75 & 76.01 & 66.94 \\
    HM-E (R) &  12.63 & 2.27 & 79.24 & 73.34 & 12.83 & \textbf{2.62} & \textbf{76.59} & \textbf{67.74} \\
    \hline
    HM-E (R) + Gib-E (F) & 12.74 & 2.17 & \textbf{81.00} & \textbf{75.44} & 13.57 & 2.65 & 76.33 & 66.97 \\
    HM-E (R) + Gib-E (R) & 12.41 & \textbf{2.02} & 80.51 & 74.88 &
    13.16 & \textbf{2.53} & \textbf{78.08} & \textbf{68.31} \\
    \hline \hline
  \end{tabular}}
\end{center}
\vspace{-10pt}
\caption{Effect of augmented image quality. \textit{(F)} denotes faulty rendered images and \textit{(R)} denotes recovered images.}
\label{tab:recover_img}
\end{table}

\paragraph{Effect of Augmented Instruction}
By comparing different speakers in Table~\ref{tab:lang_aug}, we found that a simple LSTM-based model (EnvDrop~\cite{tan2019envdrop}) trained from-scratch results in a higher Bleu-4 score~\cite{papineni2002bleu} than a fine-tuned GPT-2~\cite{radford2019gpt2}. However, we are aware that producing high-fidelity and detailed navigational instructions is a long-lasting and challenging problem~\cite{dou2022foam,fried2018speaker,magassouba2021crossmap,tan2019envdrop,wang2022less,zhou2023navgpt}, but we only experimented with two simple models. Numbers in Table~\ref{tab:lang_aug} indicate that the generated instructions are of low quality while showing a large influence on learning to navigate, implying that pairing the augmented trajectories with better instructions could be promising future work.

\begin{table}[t!]
  \begin{center}
  \resizebox{0.80\columnwidth}{!}{
  \begin{tabular}{c|c|rrrr}
    \hline \hline
      \multirow{2}{*}{Speaker} & \multicolumn{1}{c|}{Instruction Quality} &\multicolumn{4}{c}{R2R Val-Unseen} \\
     \cline{2-6} &\multicolumn{1}{c|}{Bleu-4$\uparrow$}  & \multicolumn{1}{c}{TL} & \multicolumn{1}{c}{NE$\downarrow$} & \multicolumn{1}{c}{SR$\uparrow$} &
    \multicolumn{1}{c}{SPL$\uparrow$} \Tstrut\\
    \hline \hline
    GPT-2 & 24.36 & 13.98 & 2.74 & 75.82 & 66.08  \\
    EnvDrop & \textbf{27.66} & 12.83  & \textbf{2.62} & \textbf{76.59 }& \textbf{67.74} \\
    \hline \hline
  \end{tabular}}
\end{center}
\vspace{-10pt}
\caption{Quality of generated instructions and their influence on agent's performance on the R2R dataset.} 
\label{tab:lang_aug}
\end{table}

\begin{table*}[t]
  \begin{center}
  \resizebox{1.00\textwidth}{!}{
  \begin{tabular}{c|ccc|ccc|rrrr|rrrr}
    \hline \hline
    \multicolumn{1}{c|}{\multirow{2}{*}{Method \#}} & \multicolumn{3}{c|}{Pre-training Data} & \multicolumn{3}{c|}{Fine-tuning Data} & \multicolumn{4}{c|}{R2R Val-Seen} &
    \multicolumn{4}{c}{R2R Val-Unseen} \\
    \cline{2-15} & 
    \multicolumn{1}{c}{R2R} & \multicolumn{1}{c}{PREV} & \multicolumn{1}{c|}{HM-E (ours)} &  \multicolumn{1}{c}{R2R} & \multicolumn{1}{c}{PREV} & \multicolumn{1}{c|}{HM-E (ours)} & \multicolumn{1}{c}{TL} & \multicolumn{1}{c}{NE$\downarrow$} & \multicolumn{1}{c}{SR$\uparrow$} & \multicolumn{1}{c|}{SPL$\uparrow$} & \multicolumn{1}{c}{TL} & \multicolumn{1}{c}{NE$\downarrow$} & \multicolumn{1}{c}{SR$\uparrow$} & \multicolumn{1}{c}{SPL$\uparrow$}\Tstrut\\
    \hline \hline
    1 & 
    \checkmark &  &  &
    \checkmark &  &  & 
     12.02 & 3.39 & 69.83 & 64.30 &
     12.45 & 4.04 & 65.26 & 56.91\\
    2 & 
    \checkmark & \checkmark &  & 
    \checkmark &  &  & 
     13.28 & 2.51 & 76.89 & 69.71 &
     13.53 & 3.06 & 72.92 & 62.82\\
    3 & 
    \checkmark &  & \checkmark & 
    \checkmark &  &  & 
     12.80 & 2.69 & 75.02 & 68.71 &
     12.66 & 2.79 & 74.96 & 65.90\\
     4 & 
    \checkmark & \checkmark & \checkmark & 
    \checkmark &  &  & 
     12.63 & 2.27 & 79.24 & 73.34 &
     12.83 & 2.62 & 76.59 & 67.74\\
     \hline
     5 & 
    \checkmark & \checkmark & \checkmark & 
    \checkmark & \checkmark &  & 
     12.31 & 2.20 & 80.51 & \textbf{75.75} &
     12.86 & 2.65 & 75.78 & 66.36\\
     6 & 
    \checkmark & \checkmark & \checkmark &
    \checkmark & & \checkmark & 
     13.38 & \textbf{2.12} & 80.02 & 73.52 &
     13.32 & \textbf{2.46} & \textbf{79.10} & \textbf{68.66}\\
     7 & 
    \checkmark & \checkmark & \checkmark & 
    \checkmark & \checkmark & \checkmark & 
     12.62 & 2.18 & \textbf{80.71} & 75.29 &
     13.22 & 2.58 & 77.10 & 67.23 \\
    \hline \hline
  \end{tabular}}
\end{center}
\vspace{-10pt}
\caption{Influence of applying augmented data in pre-training and fine-tuning on agent's performance.}
\vspace{-3pt}
\label{tab:pretrain_finetune}
\end{table*}

\subsection{How to Utilize Large-Scale Data?}
\label{subsec:utilize_data}

\paragraph{Data for Pre-Training and Fine-Tuning} 

Pre-training and fine-tuning are two essential stages where augmented data can directly impact. In Table~\ref{tab:pretrain_finetune}, we investigate how to effectively apply the original R2R dataset, PREV~\cite{hao2020prevalent}, and our HM-E data in the two processes.
First, comparing applying PREV and HM-E (Method\#2 and \#3) in pre-training, it is unsurprisingly that an agent benefits more from learning in environments different from downstream scenes. 
A better result of Method\#4 shows that PREV and our HM-E complement each other in the pre-training phase.
Then, we investigate the effect of applying augmented data in fine-tuning, in which the motivation is to avoid overfitting the small downstream dataset. Compare Method\#4 to Method\#5, \#6, and \#7; it is clear that it is very beneficial to keep the data augmented from the addition environments (HM-E) in fine-tuning (+2.51\% SR in Val-Unseen). Moreover, by doing so, the generalization gap between navigating in seen and unseen environments has been reduced to less than 1\% SR (80.02\% vs. 79.10\%), reflecting the importance of maintaining high visual diversity in training. Compare Method\#6 and Method\#7, including PREV in fine-tuning harms the performance likely because it will cause the learning to overfit the 61 MP3D scenes.
In addition to Table~\ref{tab:pretrain_finetune}, our experiment shows that adding Gib-E to the pre-training phrase can improve the result, while applying it in fine-tuning does not show a noticeable difference. This is likely because the generated instruction-trajectory pairs from Gibson environments have a larger gap to the Matterport scenes, which will introduce noise and is unsuitable for fine-tuning.

\paragraph{Effect of Pre-training Tasks} 
In Table~\ref{tab:pretrain_tasks}, we further investigate the effect of three proxy tasks, MLM, SAP, and MRM, on pre-training the best performing model in Table~\ref{tab:pretrain_finetune} (Method\#6). Results show that both MLM and SAP are very effective pre-training tasks that can greatly enhance the agent's performance when applied alone, and they are complementary since combining the two tasks can lead to a larger improvement (+16.49\% SPL higher than without pre-training). However, learning MRM with the other two proxy tasks slightly degenerates the results. We suspect this is because an agent can already learn very rich and generalizable semantic representations from large and diverse augmented data, whereas predicting the probability distribution of object categories for masked images introduces too much noise to the learning process.

\vspace{5pt}

Based on the findings from our experiments, we pre-train our agent with MLM and SAP on R2R, PREV, and our \ours{} datasets to get the best pre-trained model, and fine-tune on R2R and HM-E for best performance.

\begin{table}[t]
  \begin{center}
  \resizebox{\columnwidth}{!}{
  \begin{tabular}{ccc|rrrr|rrrr}
    \hline \hline
     \multicolumn{3}{c|}{Pre-training Tasks} & \multicolumn{4}{c|}{R2R Val-Seen} &\multicolumn{4}{c}{R2R Val-Unseen} \\
    \cline{1-11} MLM & SAP & MRM & \multicolumn{1}{c}{TL} & \multicolumn{1}{c}{NE$\downarrow$} & \multicolumn{1}{c}{SR$\uparrow$} &
    \multicolumn{1}{c|}{SPL$\uparrow$} & \multicolumn{1}{c}{TL} & \multicolumn{1}{c}{NE$\downarrow$} & \multicolumn{1}{c}{SR$\uparrow$} &
    \multicolumn{1}{c}{SPL$\uparrow$} \Tstrut\\
    \hline \hline
    &  & & 11.96 & 3.61 & 66.01 & 60.66 & 14.47 & 4.26 & 62.62 & 51.25 \\
    \checkmark & & & 14.13 & 2.91 & 74.93 & 65.95 & 15.75 & 3.62 & 69.14 & 57.47 \\
     & \checkmark  & & 12.20 & 1.99 & 80.61 & 75.07 & 12.89 &2.89 & 74.20 & 65.40 \\
    \checkmark  & \checkmark  &  & 12.63 & 2.27 & 79.24 & 73.34 &
    12.83 & \textbf{2.62} & \textbf{76.59} & \textbf{67.74}\\
    \checkmark  & \checkmark  & \checkmark & 12.37 & \textbf{1.97} & \textbf{81.88} & \textbf{75.72} & 13.69 & 2.73 & 75.82 & 66.62 \\
    \hline \hline
  \end{tabular}}
\end{center}
\vspace{-10pt}
\caption{Influence of pre-training tasks. MLM, SAP, and MRM denote masked language modeling, single-action prediction, and masked region modeling.}
\label{tab:pretrain_tasks}
\vspace{-5pt}
\end{table}

\begin{table}[t!]
  \begin{center}
  \resizebox{\columnwidth}{!}{
  \begin{tabular}{l|ccc|ccc}
    \hline \hline
    \multicolumn{1}{c|}{\multirow{2}{*}{Methods}}  & \multicolumn{3}{c|}{Val-Unseen} & \multicolumn{3}{c}{Test-Unseen}  \\
    \cline{2-7} & \multicolumn{1}{c}{OSR$\uparrow$} & \multicolumn{1}{c}{SR$\uparrow$} &
    \multicolumn{1}{c|}{SPL$\uparrow$} & \multicolumn{1}{c}{OSR$\uparrow$} &
    \multicolumn{1}{c}{SR$\uparrow$} &
    \multicolumn{1}{c}{SPL$\uparrow$} \Tstrut\\
    \hline \hline
    RecBERT~\cite{hong2020recurrent} & 27.66 & 25.53 & 21.06 & 26.67 & 24.62 & 19.48  \\
    SIA~\cite{lin2021scene}  & 44.67 & 31.53 & 16.28 & 44.56 &30.80 & 14.85 \\ 
    HAMT~\cite{chen2021hamt} & 36.84 & 32.95 & 30.20 & 33.41 & 30.40 & 26.67 \\ 
    DUET~\cite{chen2022duet} & 51.07 & 46.98 & 33.73 & 56.91 & 52.51 & 36.06 \\
    AutoVLN~\cite{chen2022hm3dlearning} & 62.14 & 55.89 & 40.85 & 62.30 & 55.17 & 38.88  \\
    \hline
    DUET+\ours{} (ours) & \textbf{63.85} & \textbf{56.97} & \textbf{41.84} & \textbf{62.65} & \textbf{56.13} & \textbf{39.52} \\
    \hline \hline
  \end{tabular}}
\end{center}
\vspace{-10pt}
\caption{Navigation performance on REVERIE dataset.}
\label{tab:reverie_results}
\vspace{-10pt}
\end{table}

\subsection{Evaluate on Various VLN Tasks}

\paragraph{R2R} 
Table~\ref{tab:r2r_full} compares agents' single-run performance on the R2R dataset. We can see that training DUET model with our \ours{} data results in 8\% SR and 8\% SPL absolute improvement on the test split\footnote{On the R2R test-unseen leaderboard: \url{https://eval.ai/web/challenges/challenge-page/97/leaderboard/270}, our method surpasses all single-run results and outperforms all previous models applying beam-search or pre-exploration (see \textit{Appendix}~\ref{sec_2}).}, which also greatly outperforms the previous best method BEVbert~\cite{an2022bevbert}. As suggested in MARVAL~\cite{kamath2022marval}, we also experiment with applying a more powerful visual encoder, CLIP ViT-H/14~\cite{radford2021clip}, and the image augmentation method EnvEdit~\cite{li2022envedit} to our approach, leading to a remarkable 80\% SR, and reducing the long-lasting generalization gap between seen and unseen environments~\cite{zhangdiagnosing} to less than 1\%. It is interesting to notice that the remaining gap towards human performance (6\% SR) is similar to the difference between the agent's OSR and SR (6{\textasciitilde}7\%), which suggests that it might be important for future work to improve the policy network to tackle the stopping problem given large-scale data.

\begin{table*}[ht]
\centering
\resizebox{\textwidth}{!}{\begin{tabular}{l|ccccc | ccccc|ccccc}
\hline \hline
\multicolumn{1}{c|}{\multirow{2}{*}{Methods}} & \multicolumn{5}{c|}{R2R Val-Seen} & \multicolumn{5}{c|}{R2R Val-Unseen} & \multicolumn{5}{c}{R2R Test-Unseen}\Tstrut \\
\cline{2-16}
\multicolumn{1}{c|}{} & 
\multicolumn{1}{c}{TL} & \multicolumn{1}{c}{NE$\downarrow$} & \multicolumn{1}{c}{OSR$\uparrow$} & \multicolumn{1}{c}{SR$\uparrow$} & \multicolumn{1}{c|}{SPL$\uparrow$} & 
\multicolumn{1}{c}{TL} & \multicolumn{1}{c}{NE$\downarrow$} & \multicolumn{1}{c}{OSR$\uparrow$} & \multicolumn{1}{c}{SR$\uparrow$} & \multicolumn{1}{c|}{SPL$\uparrow$} & 
\multicolumn{1}{c}{TL} & \multicolumn{1}{c}{NE$\downarrow$} & \multicolumn{1}{c}{OSR$\uparrow$} & \multicolumn{1}{c}{SR$\uparrow$} & \multicolumn{1}{c}{SPL$\uparrow$} \\
\hline \hline
Human                   
& -- & -- & -- & -- & -- 
& -- & -- & -- & -- & --
& 11.85 & 1.61 & 90 & 86 & 76 \\
\hline
Seq2Seq~\cite{anderson2018r2r}
& 11.33 & 6.01 & 53 & 39 & -- 
& 8.39 & 7.81 & 28 & 21 & -- 
& 8.13 & 7.85 & 27 & 20 & -- \\
Speaker Follower~\cite{fried2018speaker}
& -- & 3.36 & 74 & 66 & -- 
& -- & 6.62 & 45 & 36 & -- 
& 14.82 & 6.62 & - & 35 & 28  \\
RCM~\cite{wang2019reinforced}
& 10.65 & 3.53 & 75 & 67 & -- 
& 11.46 & 6.09 & 50 & 43 & -- 
& 11.97 & 6.12 & 50 & 43 & 38 \\
SSM~\cite{wang2021structured}
& 14.70 & 3.10 & 80 & 71 & 62
& 20.70 & 4.32 & 73 & 62 & 45
& 20.40 & 4.57 & 70 & 61 & 46 \\
EnvDrop~\cite{tan2019envdrop}
& 11.00 & 3.99 & -- & 62 & 59
& 10.70 & 5.22 & -- & 52 & 48
& 11.66 & 5.23 & 59 & 51 & 47 \\
PREVALENT~\cite{hao2020prevalent} $\dag$
& 10.32 & 3.67 & -- & 69 & 65
& 10.19 & 4.71 & -- & 58 & 53
& 10.51 & 5.30 & 61 & 54 & 51 \\
EntityGraph~\cite{hong2020graph}
& 10.13 & 3.47 & -- & 67 & 65
& 9.99 & 4.73 & -- & 57 & 53
& 10.29 & 4.75 & 61 & 55 & 52 \\
NvEM~\cite{an2021neighbor}
& 11.09 & 3.44 & -- & 69 & 65
& 11.83 & 4.27 & -- & 60 & 55
& 12.98 & 4.37 & 66 & 58 & 54 \\
AirBert~\cite{guhur2021airbert} $\dag$$\ddag$
& 11.09 & 2.68 & -- & 75 & 70
& 11.78 & 4.10 & -- & 62 & 56 
& 12.41 & 4.13 & -- & 62 & 57 \\
\vlnbert~\cite{hong2020recurrent} $\dag$
& 11.13 & 2.90 & -- & 72 & 68
& 12.01 & 3.93 & -- & 63 & 57
& 12.35 & 4.09 & 70 & 63 & 57 \\
MARVAL~\cite{kamath2022marval} $\dag$$\ddag$
& 10.60 & 2.99 & -- & 73 & 69
& 10.15 & 4.06 & -- & 65 & 61
& 10.22 & 4.18 & 67 & 62 & 58 \\
EnvMix~\cite{liu2021envmixup} $\dag$
& 10.88 & 2.48 & -- & 75 & 72
& 12.44 & 3.89 & -- & 64 & 58
& 13.11 & 3.87 & 72 & 65 & 59 \\
HAMT~\cite{chen2021hamt} $\dag$
& 11.15 & 2.51 & -- & 76 & 72
& 11.46 & 2.29 & -- & 66 & 61
& 12.27 & 3.93 & 72 & 65 & 60 \\
SnapEnsemble~\cite{qin2021ensemble} $\dag$$^{\circ}$
& -- & -- & -- & -- & --
& 12.05 & 3.63 & -- & 67 & 60
& 12.71 & 3.82 & -- & 65 & 60 \\
HOP+~\cite{qiao2023hop+} $\dag$
& 11.31 & 2.33 & -- & 78 & 73 
& 11.76 & 3.49 & -- & 67 & 61
& 12.67 & 3.71 & -- & 66 & 60 \\
TD-STP~\cite{zhao2022target} $\dag$
& -- & 2.34 & 83 & 77 & 73 
& -- & 3.22 & 76 & 70 & 63 
& -- & 3.73 & 72 & 67 & 61 \\
DUET~\cite{chen2022duet} $\dag$
& 12.32 & 2.28 & 86 & 79 & 73 
& 13.94 & 3.31 & 81 & 72 & 60
& 14.73 & 3.65 & 76 & 69 & 59 \\
BEVBert~\cite{an2022bevbert} $\dag$
& 13.56 & 2.17 & \blue{88} & \blue{81} & \textbf{74}
& 14.55 & 2.81 & 84 & 75 & 64
& 15.87 & 3.13 & 81 & 73 & 62 \\
\hline
DUET+\ours{} (ours) $\dag$$\ddag$
 & 11.90 & \textbf{2.16} & \textbf{87} & \textbf{80} & \blue{75}
 & 12.40 & \textbf{2.34} & \textbf{87} & \textbf{79} & \blue{70}
 & 14.27 & \textbf{2.73} & \textbf{83} & \textbf{77} & \textbf{68} \\
DUET*+\ours{} (ours) $\dag$$\ddag$
 & 13.24 & \blue{2.12} & \textbf{87} & \blue{81} & \blue{75}
 & 14.09 & \blue{2.09} & \blue{88} & \blue{81} & \blue{70}
& 13.93 & \blue{2.27} & \blue{86} & \blue{80} & \blue{70}\\
\hline\hline
\end{tabular}}
\vspace{-5pt}
\caption{Comparison of single-run performance on R2R dataset. $\dag$: Methods that apply vision-language-action pre-training. $\ddag$: Methods that use additional visual data than MP3D. $^{\circ}$: Model ensemble. *: Applying EnvEdit as image augmentation and CLIP ViT-H\/14 as image features.}
\label{tab:r2r_full}
\vspace{-13pt}
\end{table*}

\paragraph{REVERIE} 
We show in Table~\ref{tab:reverie_results} that our method achieves the new state-of-the-art results in all metrics on the REVERIE task. Our method surpasses AutoVLN, which uses all the 1000 HM3D environments for pre-training, by 0.94\% in success rate and 0.64\% in SPL on the test leaderboard with only 800 HM3D scenes and 491 Gibson low-quality environments. This again validates the effectiveness of our high-quality connectivity graphs and image recovery in our large-scale training paradigm.

\paragraph{CVDN}

As shown in Table~\ref{tab:cvdn_sota_cmpr}, our method achieves the new state-of-the-art performance on the CVDN test-unseen split, which largely improves the goal progress (GP) of the previous SoTA by 1.41 meters (a relative gain of 25.26\%). This result shows that our R2R-style augmented data can generalize to a different VLN task with a distinct type of instructions, likely because visual scarcity is the major bottleneck in learning VLN, as suggested in Table~\ref{tab:quan_env_sample}.

\paragraph{R2R-CE}

Although our augmented \ours{} data only contains discrete instruction-trajectory pairs, it can benefit the agent's performance in continuous environments with the support of the candidate waypoint predictor~\cite{hong2022bridging} (Table~\ref{tab:r2rce_results}). Compared to the previous method, which applies very strong pre-trained visual representations~\cite{hong2023ego2map,wang2022internvideo}, our method still demonstrates obvious improvement, which reflects the effectiveness of generating in-domain data for learning VLN.

\begin{table}[t!]
  \begin{center}
  \resizebox{0.85\columnwidth}{!}{
\begin{tabular}{l|c|c|c} 
\hline \hline
 \multicolumn{1}{c|}{\multirow{2}{*}{Methods}} & \multicolumn{1}{c|}{Val-Seen} & \multicolumn{1}{c|}{Val-Unseen} & \multicolumn{1}{c}{Test-Unseen} \\
 
 \cline{2-4}& \multicolumn{1}{c|}{GP$\uparrow$} & \multicolumn{1}{c|}{GP$\uparrow$} & \multicolumn{1}{c}{GP$\uparrow$} \\
 \hline \hline
PREVALENT \cite{hao2020prevalent} & -- & 3.15 & 2.44 \\
MT-RCM+EnvAg \cite{wang2020environment} & 5.07 & 4.65 & 3.91 \\
NDH-Full~\cite{kim2021ndh} & - & 5.51 & 5.27 \\
HAMT~\cite{chen2021hamt} & 6.91 & 5.13 & 5.58  \\
MTVM~\cite{lin2022multimodal} & -- & 5.15 & 4.82 \\
\hline
DUET+\ours{} (Ours) & \textbf{8.13} & \textbf{6.12} & \textbf{6.97}  \\
\hline \hline
\end{tabular}}
\end{center}
\vspace{-10pt}
\caption{Navigation performance on CVDN dataset.}
\label{tab:cvdn_sota_cmpr}
\vspace{-9pt}
\end{table}

\begin{table}[t!]
  \begin{center}
  \resizebox{\columnwidth}{!}{
\begin{tabular}{l|cccc|ccc}
    \hline \hline
    \multicolumn{1}{c|}{\multirow{2}{*}{Methods}} &  
    \multicolumn{4}{c|}{R2R-CE Val-Unseen} &
    \multicolumn{3}{c}{R2R-CE Test-Unseen} \\
    \cline{2-8} & \multicolumn{1}{c}{NE$\downarrow$} &
    \multicolumn{1}{c}{nDTW$\uparrow$} &
     \multicolumn{1}{c}{SR$\uparrow$} & \multicolumn{1}{c|}{SPL$\uparrow$} & \multicolumn{1}{c}{NE$\downarrow$} & \multicolumn{1}{c}{SR$\uparrow$} & \multicolumn{1}{c}{SPL$\uparrow$} \Tstrut\\
    \hline \hline
    CMA~\cite{krantz2020navgraph}  & 7.37 & 40 & 32 & 30 & 7.91 & 28 & 25 \\
    LAW~\cite{raychaudhuri2021law} & -- & -- & 35 & 31 & -- & --  &  -- \\
    Waypoint Models~\cite{krantz2021waypoint}  & 6.31 & --  & 36 & 34  & 6.65  & 32 & 30 \\
    WS-MGMap~\cite{chen2022weakly} & 6.28 & - & 39 & 34 & 7.11 & 35 & 28 \\
    \vlnbert$\dag$~\cite{hong2022bridging}  & 5.74 & 53 & 44 & 39 & 5.89 & 42 & 36 \\
    Sim2Sim~\cite{krantz2022sim} & 6.07 & - & 43 & 36 & 6.17 & 44 & 37 \\ 
    \vlnbert+Ego$^2$-Map$\dag$~\cite{hong2023ego2map} & 4.94 & 60 & 52 & 46 & 5.54 & 47 & 41 \\     HAMT+InternVideo$\dag$~\cite{wang2022internvideo}  & 4.95 & 62 & 53 &  48 & -- & -- & -- \\
    \hline 
    HAMT+\ours{}$\dag$ (ours) & \textbf{4.80} & \textbf{64} & \textbf{55} & \textbf{51} & \textbf{5.11} & \textbf{55} & \textbf{50}  \\
    \hline \hline
  \end{tabular}}
\end{center}
\vspace{-10pt}
\caption{Navigation performance on the R2R-CE datasets. $\dag$: Methods that applies candidate waypoint predictor~\cite{hong2022bridging} to support high-level action space. }
\label{tab:r2rce_results}
\vspace{-10pt}
\end{table}

\section{Conclusion}
\label{sec:conclude}
\vspace{-6pt}
In this paper, we introduce a simple but effective large-scale data generation paradigm for learning vision-and-language navigation (\ours{}). The method applies thousands of photo-realistic environments from HM3D and Gibson datasets, and creates millions of instruction-trajectory pairs for training. Apart from the unsurprising improvement of learning from abundant visual data in agent performance, we demonstrate the effectiveness of building high-quality navigation graphs and using camera-quality images through comprehensive experiments. Moreover, we investigate how to properly utilize the augmented data in pre-training and fine-tuning an agent, as well as the influence of different pre-training tasks on the downstream navigation results. By following our findings as data augmentation and agent training guidelines, we achieve new state-of-the-art results on several VLN benchmarking datasets that cover distinct styles of instructions (R2R, REVERIE, CVDN) and action spaces (R2R-CE). We believe our \ours{} paradigm can be easily applied as a tool to facilitate data augmentation for VLN and other visual navigation problems, and the experiments presented in the paper can provide useful insights for future research in creating and utilizing large-scale data.

\vspace{-8pt}
\section{Acknowlegement}
\vspace{-6pt}
We thank ICCV reviewers for their helpful suggestions. This work is partially supported by the National Key R\&D Program of China (NO. 2022ZD0160100), and in part by the Shanghai Committee of Science and Technology (Grant No. 21DZ1100100).
We warmly thank Taesung Park for helpful suggestions on recovering rendered images.


{\small
\bibliographystyle{ieee_fullname}
\bibliography{egbib}
}

\clearpage
\appendix
\section*{Appendices}

We first describe the implementation details of our experiments in Sec.~\ref{sec_1}, including pre-training objectives and details of REVERIE experiments. In Sec.~\ref{sec_2}, we provide additional experiments about the effects of visual encoders, model initialization, and adding depth features. We then discuss the impact of ScaleVLN on different VLN agents and on learning the long-horizon VLN task (R4R). Leaderboard results of R2R and object grounding results for REVERIE are also included. Sec.~\ref{sec_3} and Sec.~\ref{sec_4} visualize our navigability graphs and the recovered images from Co-Modulated GAN~\cite{zhao2021comodgan}.

\section{Implementation Details (\S4\protect\footnote{Link to Section 4 in Main Paper.})}

\label{sec_1}

\subsection{Pre-Training Objectives \texorpdfstring{($\boldsymbol{\S}$}~4.1)}

 We mainly employ three proxy tasks, MLM, MRM, and SAP, for pre-training the agent. Here we describe these proxy tasks in detail. The inputs for these tasks are instruction $\mathcal{W}$ and demonstration path $\mathcal{P}$. During training, we randomly sample one task for each iteration with equal probability.

\paragraph{Masked Language Modeling (MLM)} involves predicting masked words based on textual context and the full trajectory. A special \verb|[mask]| token is used to randomly mask out 15\% of the tokens in $\mathcal{W}$. We predict the masked word distribution $p (w_i|\mathcal{W}_{\backslash i}, \mathcal{P})=f_{\text{MLM}}(x'_i)$ through a two-layer fully-connected network, where $\mathcal{W}_{\backslash i}$ is the masked instruction and $x'_i$ is the output embedding of the masked word $w_i$. The objective is to minimize the negative log-likelihood of predicting the original words: $\mathcal{L}_{\text{MLM}} = - \mathrm{log}\ p (w_i|\mathcal{W}_{\backslash i}, \mathcal{P})$. 

\paragraph{Masked Region Modeling (MRM)} is to predict labels for masked regions in history observations based on instructions and neighboring regions.  To achieve this, we randomly remove view images in $\mathcal{P}$ with a 15\% probability. For view images, the target labels are determined by an image classification model~\cite{dosovitskiy2020vit} pre-trained on ImageNet. To predict semantic labels for each masked visual token, we use a two-layer fully-connected network. The objective is to minimize the KL-divergence between the predicted and target probability distribution. 

\paragraph{Single Action Prediction (SAP)} aims to predict the next action based on the instruction and the given path. Following \cite{chen2022duet}, we predict the probability for each candidate action in the action space via a two-layer fully-connected network. The objective is to minimize the negative log probability of the target view action $\mathcal{L}_{\mathrm{SAP}} = -\mathrm{log}\ p_t(a^*_t | \mathcal{W}, \mathcal{P}_{<t})$. 

\subsection{Implementation Details of REVERIE \texorpdfstring{($\boldsymbol{\S}$}~4.1)}


REVERIE data contains trajectories that lead to target objects specified by high-level instructions. Following AutoVLN~\cite{chen2022hm3dlearning}, for every visible object at a viewpoint, we sample paths with an edge length between 4 and 9 that end at the viewpoint. We filter out objects that are more than 3 meters away from the central of the viewpoint, resulting in 518,233 paths from HM3D, and 311,976 paths from the Gibson environments.
To generate instructions in REVERIE-style, we modify the GPT-2 architecture used in AutoVLN~\cite{chen2022hm3dlearning} by only encoding the target object in the final viewpoint as the prompt to generate the instructions. Our large-scale data augmentation paradigm creates 830,209 instruction-trajectory pairs for training. This size is $\times$38 larger than the original REVERIE dataset, and $\times$3.81 larger than the augmented dataset in AutoVLN~\cite{chen2022hm3dlearning}.

We follow DUET and SIA~\cite{lin2021sia} to pre-train the model with an additional Object Grounding (OG) task, which requires selecting a target from object candidates based on high-level instruction and observations along the path. We use CLIP ViT-H/14~\cite{radford2021clip} to extract the image features, and ViT-B/16~\cite{dosovitskiy2020vit} pre-trained on ImageNet to extract the object features.
We pre-train DUET for 100k iterations with a batch size of 128 and a learning rate of $5\times 10^{-5}$ on both HM3D and Gibson environments. We compare three model checkpoints at 30k, 40k, and 50k and pick the one with the highest fine-tuning performance. Then we fine-tune DUET for 150k iterations, with batch size 32 and learning rate $2\times 10^{-5}$ on a single NVIDIA A100 GPU.

\section{Additional Experiments (\S4)}
\label{sec_2}

Here we provide additional experiments to investigate the effect of visual encoder, model initialization, and depth features. We also experiment with different model architectures (\textit{i.e.}, HAMT~\cite{chen2021hamt}) on R2R dataset, and show object grounding results for the REVERIE task.

\subsection{Effect of Visual Encoders (\texorpdfstring{$\boldsymbol{\S}$}~4.2)}

We study the effect of visual encoders in Table \ref{tab:visual_encoders.}. Here we adopt CLIP's ViT backbone with different model sizes and input patches (\textit{i.e.}, Base/16, Large/14, and Huge/14).
We can see that the vision encoder has a major influence on SPL, suggesting the agent can make fewer wrong steps and is capable of efficient navigation.

\begin{table}[h]
  \begin{center}
  \resizebox{\columnwidth}{!}{
  \begin{tabular}{l|cccc|rrrr}
    \hline \hline
     \multirow{2}{*}{Visual Encoders} & \multicolumn{4}{c|}{R2R Val-Seen} &\multicolumn{4}{c}{R2R Val-Unseen} \\
     \cline{2-9} & \multicolumn{1}{c}{TL}& \multicolumn{1}{c}{NE$\downarrow$} & \multicolumn{1}{c}{SR$\uparrow$} &
    \multicolumn{1}{c|}{SPL$\uparrow$} & \multicolumn{1}{c}{TL}& \multicolumn{1}{c}{NE$\downarrow$} & \multicolumn{1}{c}{SR$\uparrow$} &
    \multicolumn{1}{c}{SPL$\uparrow$} \Tstrut\\
    \hline \hline
    CLIP-ViT-B/16 &    12.41 & \textbf{2.02} & 80.51 & 74.88 &
      13.16 & 2.53 & 78.08 & 68.31 \\
    CLIP-ViT-L/14 & 12.62 & 2.16 & 80.04 & 74.06 & 13.13 & 2.50 & 78.08 & 68.97 \\
    CLIP-ViT-H/14 & 12.53 & 2.15 & \textbf{81.19} & \textbf{76.83} & 12.61 & \textbf{2.49} & \textbf{78.20} & \textbf{69.71} \\
    \hline \hline
  \end{tabular}}
\end{center}
\vspace{-5pt}
\caption{ Effect of visual encoders.}
\label{tab:visual_encoders.}
\end{table}

\subsection{Effect of Initialization (\texorpdfstring{$\boldsymbol{\S}$}~4.2)}

Table \ref{tab:language_init} presents the performance of initializing the navigation agent with different pre-trained models in pre-training. We discovered that utilizing BERT to initialize the language encoder does not enhance downstream performance, and even harms the performance on the validation unseen set. We attribute this to the vast domain gap between uni-modal BERT's language representations and CLIP's visual representation. Results could be improved by initializing the model with LXMERT's language encoder~\cite{tan2019lxmert}, and even more by utilizing both the language encoder and cross-modal encoder from LXMERT, indicating that incorporating pre-trained vision-and-language models could benefit agent performance.

\begin{table}[h]
  \begin{center}
  \resizebox{\columnwidth}{!}{
  \begin{tabular}{l|cccc|rrrr}
    \hline \hline
     \multirow{2}{*}{\makecell{Language Encoder \\ Initialization}} & \multicolumn{4}{c|}{R2R Val-Seen} &\multicolumn{4}{c}{R2R Val-Unseen} \\
     \cline{2-9} &  \multicolumn{1}{c}{TL}&\multicolumn{1}{c}{NE$\downarrow$} & \multicolumn{1}{c}{SR$\uparrow$} &
    \multicolumn{1}{c|}{SPL$\uparrow$} &  \multicolumn{1}{c}{TL}&\multicolumn{1}{c}{NE$\downarrow$} & \multicolumn{1}{c}{SR$\uparrow$} &
    \multicolumn{1}{c}{SPL$\uparrow$} \Tstrut\\
    \hline \hline
    Random & 12.87 & 2.29  & 78.75 & 72.61  & 12.69 & 2.72 & 75.65 &67.00  \\
    BERT & 12.43 &2.29 & 79.04 & 73.72 & 12.95 &2.76 & 75.01 & 66.57 \\
    LXMERT (lang.) & 11.73 & \textbf{2.07} & \textbf{80.22} & \textbf{75.65} & 13.17 & 2.67 & 75.86 & 67.36 \\
    LXMERT (lang.+cross.) & 12.63 & 2.27 & 79.24 & 73.34 &
    12.83 & \textbf{2.62} & \textbf{76.59} & \textbf{67.74} \\
    \hline \hline
  \end{tabular}}
\end{center}
\vspace{-5pt}
\caption{Effect of different initialization, where \textit{LXMERT (lang.)} means only initialize the language encoder with LXMERT, and \textit{LXMERT (lang.+cross.)} means initialize both the langauge encoder and cross modal encoder with LXMERT. }
\label{tab:language_init}
\end{table}

\subsection{Effect of Depth Modality (\texorpdfstring{$\boldsymbol{\S}$}~4.2)}

We also explored leveraging depth information to improve visual representations as described in Table \ref{tab:depth}. In line with previous methods such as \cite{krantz2020navgraph,krantz2021waypoint,hong2022bridging,an20221st}, we directly concatenate the depth features from DDPPO~\cite{wijmans2020ddppo} (a ResNet backbone pre-trained on PointGoal navigation with depth inputs) and the RGB features (from CLIP ViT-B/16) to create the visual representations. Our findings indicate that when not using HM3D as the augmented environment, the agent's SR is significantly better if learning from the additional depth input. However, this conclusion changes when HM3D environments are involved: the agent's SR with RGBD was slightly lower than with RGB-only. We suspect that as the data is scaled up with more visual observations and language instructions, the agent may not require additional depth information to assist decision-making.

\begin{table}[h]
  \begin{center}
  \resizebox{\columnwidth}{!}{
  \begin{tabular}{c|c|rrrr|rrrr}
    \hline \hline
     \multicolumn{1}{c|}{\multirow{2}{*}{HM3D Aug}} & \multicolumn{1}{c|}{\multirow{2}{*}{Sensor}} & \multicolumn{4}{c|}{R2R Val-Seen} &\multicolumn{4}{c}{R2R Val-Unseen} \\
     \cline{3-10} &  & \multicolumn{1}{c}{TL} & \multicolumn{1}{c}{NE$\downarrow$} & \multicolumn{1}{c}{SR$\uparrow$} &
    \multicolumn{1}{c|}{SPL$\uparrow$} & \multicolumn{1}{c}{TL} & \multicolumn{1}{c}{NE$\downarrow$} & \multicolumn{1}{c}{SR$\uparrow$} &
    \multicolumn{1}{c}{SPL$\uparrow$} \Tstrut\\
    \hline \hline
    \multirow{2}{*}{$\times$} & RGB & 13.28 & \textbf{2.51} & 76.89 & 69.71 &
     13.53 & 3.06 & 72.92 & \textbf{62.82} \\
     & RGBD & 14.16 & 2.54 & \textbf{77.18} & \textbf{69.76} &15.14 & \textbf{3.02} & \textbf{74.12} & 62.54 \\
    \hline
    \multirow{2}{*}{\checkmark} & RGB  & 12.63 & 2.27 & 79.24 & 73.34 &
     12.83 & \textbf{2.62} & \textbf{76.59} & 67.74 \\
     & RGBD & 11.24 & \textbf{2.12} & \textbf{79.73} & \textbf{75.45} & 12.93 & 2.63 & 76.46 & \textbf{68.52} \\
    \hline \hline
  \end{tabular}}
\end{center}
\vspace{-5pt}
\caption{ Effect of adding depth modality.}
\label{tab:depth}
\end{table}

\subsection{ScaleVLN with Different VLN Models (\texorpdfstring{$\boldsymbol{\S}$}~4.2)}

To evaluate the generalization ability of our \ours{} dataset, we also apply the augmented data to train different VLN agents, including Seq2Seq~\cite{anderson2018r2r}, EnvDrop~\cite{tan2019envdrop}, and HAMT \cite{chen2021hamt}. 
The HAMT model is pre-trained and fine-tuned with the same data and configurations as we pre-trained the DUET model, while we follow similar configurations of Seq2Seq and Envdrop to the original papers. All three agents are trained with the CLIP ViT-B-16 feature. The results are shown in Table \ref{tab:scalevln+x_results}. Compared to using only PREVALENT~\cite{hao2020prevalent} for augmentation, All three models significantly benefit from incorporating the ScaleVLN dataset, with 12.2\%, 3.8\%, 5.5\% 
absolute increase in SR for Seq2Seq, EnvDrop, and HAMT, respectively. This shows that
ScaleVLN strengthens models’ generalization ability.
Note that Seq2Seq and Envdrop perform better on Val-Seen when using PREVALENT, mainly caused by overfitting the training environments.

\begin{table}[h]
  \begin{center}
  \resizebox{\columnwidth}{!}{
  \begin{tabular}{l|c|c|rrr|rrr}
    \hline \hline
     \multicolumn{1}{c|}{\multirow{2}{*}{Model}} &\multicolumn{1}{c|}{\multirow{2}{*}{Pre-Train Data}} & \multicolumn{1}{c|}{\multirow{2}{*}{\makecell{Fine-Tune Data}}} & \multicolumn{3}{c|}{R2R Val-Seen} &\multicolumn{3}{c}{R2R Val-Unseen} \\
     \cline{4-9} & & & \multicolumn{1}{c}{NE$\downarrow$} & \multicolumn{1}{c}{SR$\uparrow$} &
    \multicolumn{1}{c|}{SPL$\uparrow$} & \multicolumn{1}{c}{NE$\downarrow$} & \multicolumn{1}{c}{SR$\uparrow$} &
    \multicolumn{1}{c}{SPL$\uparrow$} \Tstrut\\
    \hline \hline
         \multicolumn{1}{l|}{\multirow{2}{*}{Seq2Seq~\cite{anderson2018r2r}}}   &   - &     R2R, PREV    & \textbf{3.89} & \textbf{58.18} & \textbf{38.49} & 6.32 & 37.34 & 23.21 \\
    & - &     R2R, ScaleVLN  & 4.78 & 49.85 & 36.32  & \textbf{5.20} & \textbf{47.51} & \textbf{34.81} \\
         \hline
         \multicolumn{1}{l|}{\multirow{2}{*}{Envdrop~\cite{tan2019envdrop}}}   &    - &     R2R, PREV   & \textbf{3.65} & \textbf{66.12} & \textbf{61.72} & 4.41 & 59.22 & 52.35 \\
    & - &     R2R, ScaleVLN & 3.70 & 65.23 & 59.06 & \textbf{3.99} & \textbf{63.01} & \textbf{54.93} \\
    \hline
    \multicolumn{1}{l|}{\multirow{3}{*}{HAMT~\cite{chen2021hamt}}}   &    R2R, PREV &     R2R, PREV   & 2.58 & 74.93 & 71.52 & 3.69 & 64.90 & 60.11\\
    & R2R, PREV, ScaleVLN &     R2R & \textbf{2.15} & \textbf{79.53} & \textbf{76.64} & 3.43 & 67.56 & 62.32\\
     & R2R, PREV, ScaleVLN & R2R, ScaleVLN &  2.43 & 76.40 & 73.30 
     & \textbf{3.07} & \textbf{70.46} & \textbf{65.12} \\
    \hline \hline
  \end{tabular}}
\end{center}
\vspace{-5pt}
\caption{Influence of ScaleVLN on different VLN models.}
\label{tab:scalevln+x_results}
\end{table}

\subsection{ScaleVLN for Long-Horizon VLN (\texorpdfstring{$\boldsymbol{\S}$}~4.2)} 

We evaluate the impact of our dataset on a long-horizon VLN dataset, R4R~\cite{jain2019stay}. R4R extends the R2R dataset by concatenating two adjacent trajectories in R2R, resulting in longer navigation trajectories not biased by the shortest path prior. We directly fine-tune our pre-trained HAMT models from Table \ref{tab:scalevln+x_results} on R4R. Compared to pre-training with only R2R and PREVALENT, adding our \ours{} dataset in the pre-training stage leads to a  consistent gain, yielding +2.7\% SR, +1.5\% nDTW and +2.7\% SDTW~\cite{ilharco2019ndtw}. As suggested by the large improvement in nDTW between the ground-truth path and the executed path, our \ours{} data not only facilitate the model to reach the target but also follow the path described by the given instruction.

\begin{table}[h]
  \begin{center}
  \resizebox{\columnwidth}{!}{
  \begin{tabular}{c|c|rrrrr}
    \hline \hline
     \multicolumn{1}{c|}{\multirow{2}{*}{Pre-Train Data}} & \multicolumn{1}{c|}{\multirow{2}{*}{\makecell{Fine-Tune Data}}} &\multicolumn{5}{c}{R4R Val-Unseen} \\
     \cline{3-7}  & & \multicolumn{1}{c}{NE$\downarrow$} & \multicolumn{1}{c}{SR$\uparrow$} &
    \multicolumn{1}{c}{CLS$\downarrow$} & \multicolumn{1}{c}{NDTW$\uparrow$} &
    \multicolumn{1}{c}{SDTW$\uparrow$} \Tstrut\\
    \hline \hline
          R2R, PREV &     R4R   & 6.19 & 41.52 & 57.89 & 51.21 & 30.00  \\
    R2R, PREV, ScaleVLN &     R4R  & \textbf{6.09} & \textbf{44.20} & \textbf{59.55} & \textbf{52.77} & \textbf{32.73}  \\ 
    \hline \hline
  \end{tabular}
  }
\end{center}
\vspace{-5pt}
\caption{Effect of \ours{} on learning R4R.}
\label{tab:hamt_results}
\end{table}

\subsection{Leaderboard Results of R2R (\texorpdfstring{$\boldsymbol{\S}$}~4.4)}

We report the top seven submissions on the test-unseen leaderboard of R2R\footnote{R2R test server: \url{https://eval.ai/web/challenges/challenge-page/97/leaderboard/270}.} (Table~\ref{tab:leaderboard}). When ranking with success rate, we can see that (a) most methods have extremely low SPL (1\%) due to using beam search to find the optimal paths. Even so, our single-run result (\textit{EarlyToBed}) outperforms them by a large margin. When ranking with SPL (b), some methods pre-explored the test environments but their results are still much worse than ours. Apart from human followers, we are currently ranked first on the leaderboard.

\begin{table}[h]
\centering
\begin{minipage}{0.49\linewidth}
    \centering
    \resizebox{\textwidth}{!}{
        \begin{tabular}{lrrr}
         \hline \hline
        \multicolumn{1}{c|}{Team} & \multicolumn{1}{c}{NE$\downarrow$} & \multicolumn{1}{c}{SR$\uparrow$} & \multicolumn{1}{c}{SPL$\uparrow$} \\
            \hline \hline
            \multicolumn{1}{l|}{human} & 1.61 & 86 & 76 \\
            \hline
            \multicolumn{1}{l|}{\textbf{EarlyToBed} (ours)} & 2.27 & 80 & 70 \\
            \multicolumn{1}{l|}{LILY$^{\circ}$} & 2.54 & 78 & 1 \\
            \multicolumn{1}{l|}{Airbert$^{\circ}$} & 2.50 & 78 & 1 \\
            \multicolumn{1}{l|}{Shortest-Path-Prior$^{\circ}$} & 3.55 & 74 & 1 \\
            \multicolumn{1}{l|}{UU\_77} & 3.00 & 74 & 63 \\            
            \multicolumn{1}{l|}{TAIIC$^{\circ}$} & 2.99 & 74 & 1 \\
            \hline \hline
        \end{tabular}   
    }
    \small (a) Top 7 in SR.
\end{minipage}
\begin{minipage}{0.49\linewidth}
    \centering
    \resizebox{\textwidth}{!}{
        \begin{tabular}{lrrr}
         \hline \hline
        \multicolumn{1}{c|}{Team} & \multicolumn{1}{c}{NE$\downarrow$} & \multicolumn{1}{c}{SR$\uparrow$} & \multicolumn{1}{c}{SPL$\uparrow$} \\
            \hline \hline
            \multicolumn{1}{l|}{human} & 1.61 & 86 & 76 \\
            \hline
            \multicolumn{1}{l|}{\textbf{EarlyToBed} (ours)} & 2.27 & 80 & 70 \\
            \multicolumn{1}{l|}{TAIICX$\dag$} & 3.00 & 73 & 69 \\
            \multicolumn{1}{l|}{Active Exploration$\dag$} & 3.30 & 70 & 68 \\
            \multicolumn{1}{l|}{sponge} & 3.26 & 71 & 67 \\ 
            \multicolumn{1}{l|}{Auxiliary Reasoning$\dag$} & 3.96 & 68 & 65 \\
            \multicolumn{1}{l|}{SE-Mixed} & 3.52 & 70 & 65 \\       
            \hline \hline
        \end{tabular}
    }
    \small (b) Top 7 in SPL.
\end{minipage}
\vspace{-0.2cm}
\caption{R2R leaderboard results (28.JUL.2023). $^{\circ}$: Beam search. $\dag$: Pre-exploration.}
\label{tab:leaderboard}
\end{table}

\subsection{REVERIE Object Grounding Result (\texorpdfstring{$\boldsymbol{\S}$}~4.4)}

We report the success rate of remote object grounding (RGS) and its path length-weighted result (RGSPL). As shown in Table~\ref{tab:reverie_obj_grounding}, \ours{} achieves state-of-the-art performance on object grounding task on the test leaderboard, comparable to the previous best method AutoVLN~\cite{chen2022hm3dlearning}.

\begin{table}[h]
  \begin{center}
  \resizebox{\columnwidth}{!}{
  \begin{tabular}{l|rrrr|rrrr}
    \hline \hline
     \multicolumn{1}{c|}{\multirow{2}{*}{Models}} & \multicolumn{4}{c|}{REVERIE Val-Unseen} &\multicolumn{4}{c}{REVERIE Test-Unseen} \\
     \cline{2-9} & \multicolumn{1}{c}{SR$\uparrow$}& \multicolumn{1}{c}{SPL$\uparrow$} & \multicolumn{1}{c}{RGS$\uparrow$} &
    \multicolumn{1}{c|}{RGSPL$\uparrow$} & \multicolumn{1}{c}{SR$\uparrow$}& \multicolumn{1}{c}{SPL$\uparrow$} & \multicolumn{1}{c}{RGS$\uparrow$} &
    \multicolumn{1}{c}{RGSPL$\uparrow$} \Tstrut\\
    \hline \hline
    SIA~\cite{lin2021sia} & 31.53 & 16.28 & 22.41 & 11.56 & 30.80 & 14.85 & 19.02 & 9.20 \\ 
    HAMT~\cite{chen2021hamt} & 32.95 & 30.20 & 18.92 & 17.28 & 30.40 & 26.67 & 14.88 & 13.08 \\ 
    DUET~\cite{chen2022duet} & 46.98 & 33.73 & 32.15 & 23.03 & 52.51 & 36.06 & 31.88 & 22.06 \\ 
    AutoVLN~\cite{chen2022hm3dlearning} & 55.89 & 40.85 & \textbf{36.58} & \textbf{26.76} & 55.17 & 38.88 & 32.23 & 22.68 \\ 
    \hline
    DUET+\ours{}(ours) & \textbf{56.97} & \textbf{41.84} & 35.76 & 26.05 & \textbf{56.13} & \textbf{39.52} & \textbf{32.53} & \textbf{22.78} \\ 
    \hline \hline
  \end{tabular}}
\end{center}
\vspace{-5pt}
\caption{Object grounding performance on REVERIE.}
\label{tab:reverie_obj_grounding}
\end{table}

\section{Comparison of Navigability Graphs (\S3.2)}
\label{sec_3}

\begin{figure*}[t]
    \centering
    \includegraphics[width=0.92\textwidth]{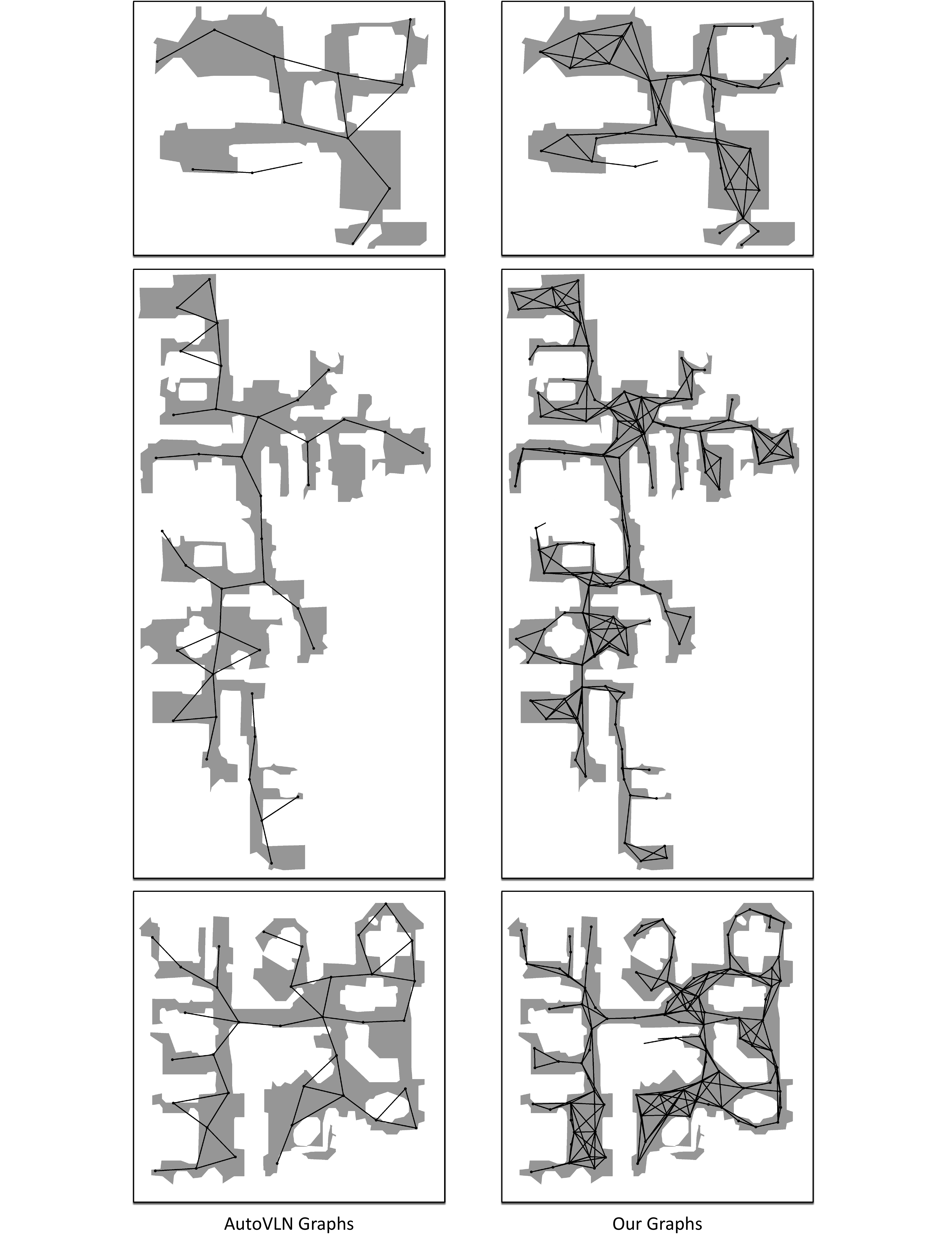}
    \caption{Comparison of navigability graphs between AutoVLN~\cite{chen2022hm3dlearning} and our \ours{}.}
    \label{fig:graph_com}
\end{figure*}

We visualize the navigability graphs produced by AutoVLN~\cite{chen2022hm3dlearning} and our method for several HM3D environments in Figure~\ref{fig:graph_com}. We can see that our graphs are denser, covering more regions, have viewpoints away from obstacles, and are fully traversable in open space.

\section{Recover High Quality Images (\S3.2)}
\label{sec_4}

As introduced in Main Paper \S3.2, we apply the Co-Modulated GAN~\cite{zhao2021comodgan} to recover the corrupted images rendered from the HM3D and Gibson environments. Specifically, we first render a panorama of shape 512$\times$1024 from the 3D mesh at each viewpoint. Then, we crop four images of shape 512$\times$512 centered at 0$^{\circ}$, 90$^{\circ}$, 180$^{\circ}$ and 270$^{\circ}$ of the panorama (with overlapping), and recover them separately. Note that, in VLN, the panoramic observation at a viewpoint is represented by 36 single-view images at 12 viewing angles and three elevations~\cite{anderson2018r2r}. We directly extract their corresponding regions from the four recovered images to obtain these single-view images for pre-training an agent.

Table~\ref{tab:recovered_envs} visualizes the difference between the rendered images and our recovered images. 
First, we can see that our method can recover missing regions, including outdoor scenes such as sky and trees (Example 1 \& 4) and indoor scenes such as floor and walls (Example 6).
Besides, the recovered images usually have less blurry or distorted areas, and the object boundaries are much clearer and sharper. For instance, the ceiling light in Example 2, the chairs in Example 3, and the door frames in Example 5.
Even for the highly corrupted images from Gibson (Examples 4--6), we can see that the method can still recover the scene to a reasonable quality.

\begin{table*}[t]
    \centering
    \resizebox{\textwidth}{!}{
    \begin{tabular}{cccc}
\hline \hline
Examples & 
Environments & 
\multicolumn{1}{c}{Rendered} &
\multicolumn{1}{c}{Recovered} \\ \hline \hline
 1 & HM3D &  \begin{minipage}{0.8\columnwidth}
   \includegraphics[width=\columnwidth]{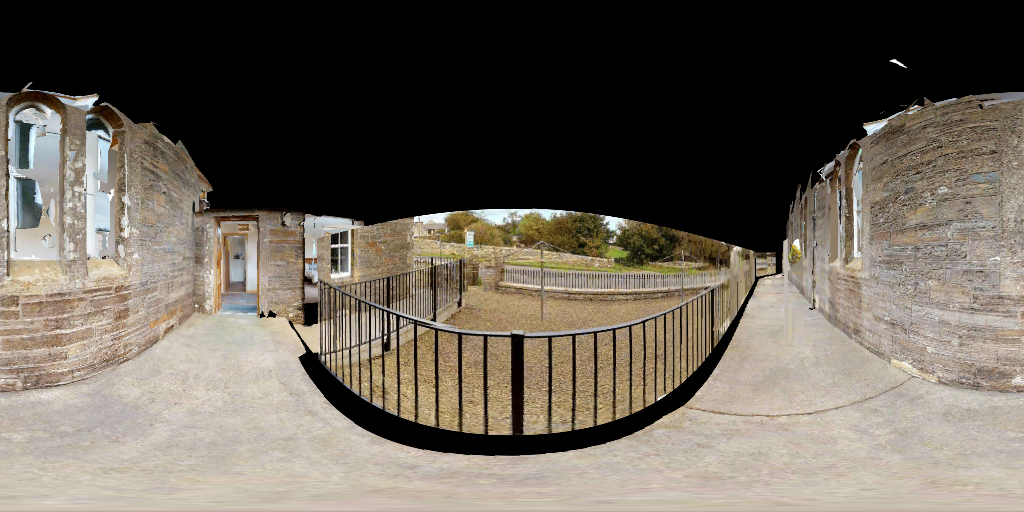} 
    \end{minipage} & \begin{minipage}{0.8\columnwidth}
      \includegraphics[width=\columnwidth]{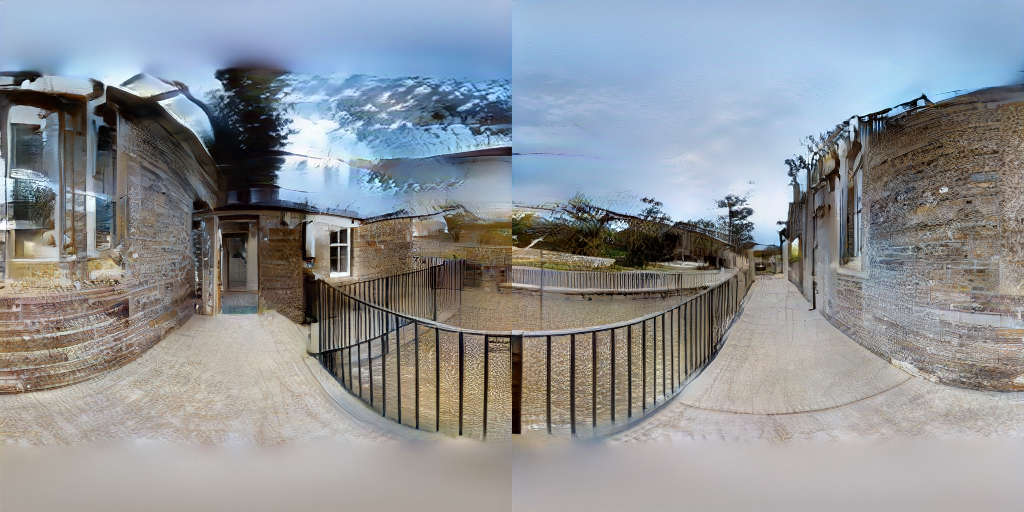} 
    \end{minipage}  \\
    \hline
2 &    HM3D &  \begin{minipage}{0.8\columnwidth}
   \includegraphics[width=\columnwidth]{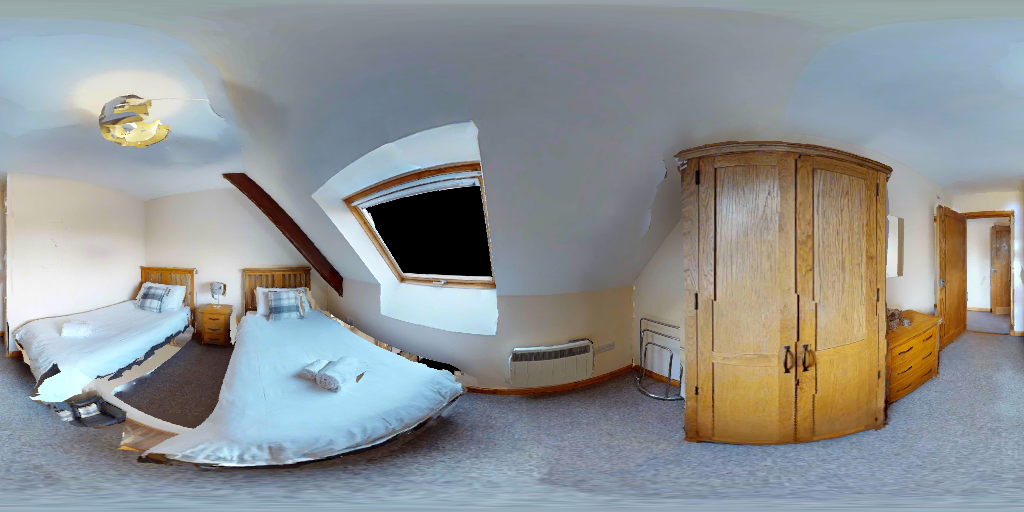} 
    \end{minipage} & \begin{minipage}{0.8\columnwidth}
      \includegraphics[width=\columnwidth]{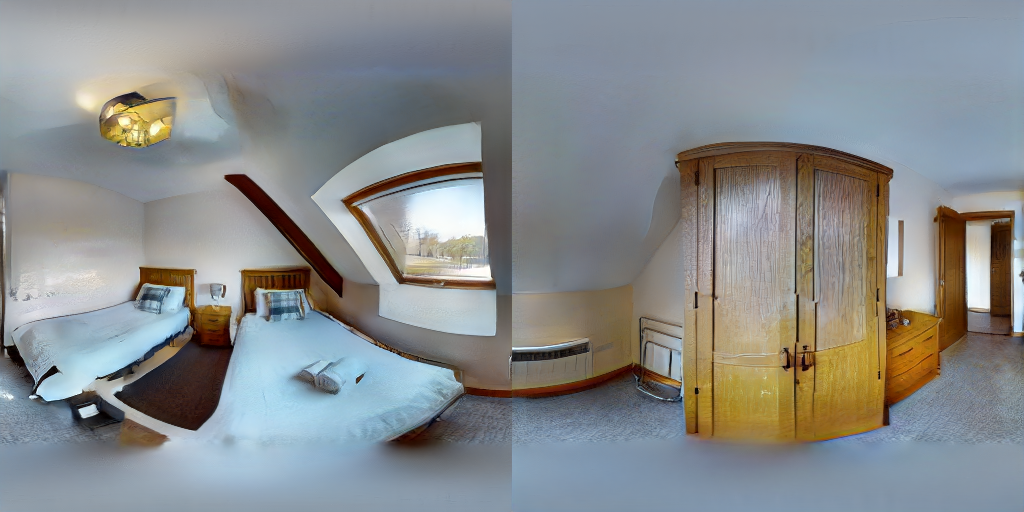} 
    \end{minipage}  \\
    \hline
 3 &   HM3D &  \begin{minipage}{0.8\columnwidth}
   \includegraphics[width=\columnwidth]{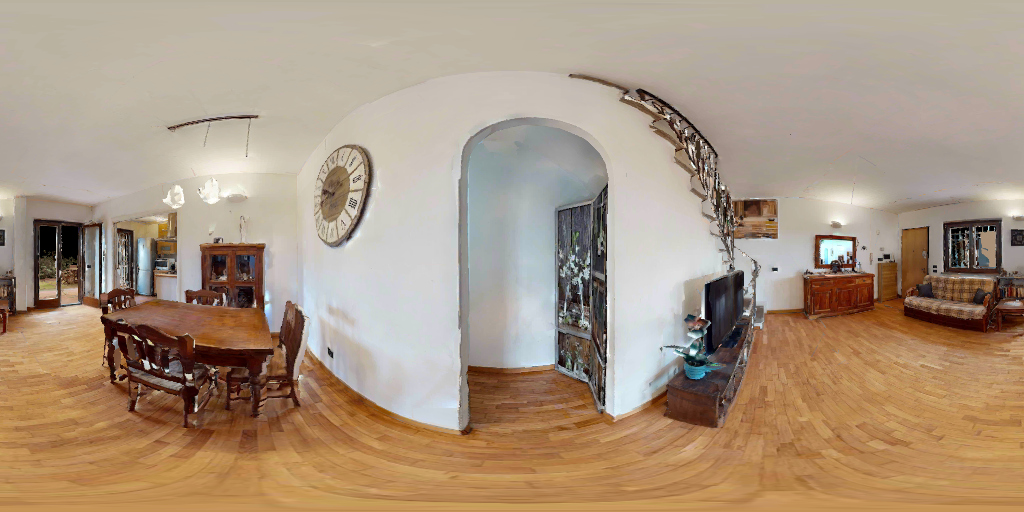} 
    \end{minipage} & \begin{minipage}{0.8\columnwidth}
      \includegraphics[width=\columnwidth]{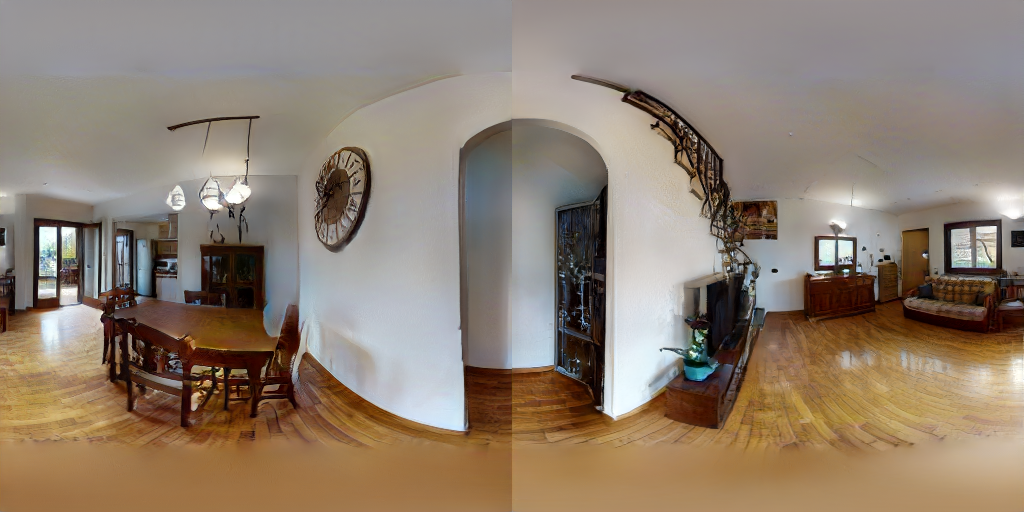} 
    \end{minipage}  \\
    \hline
    4 &   Gibson &  \begin{minipage}{0.8\columnwidth}
   \includegraphics[width=\columnwidth]{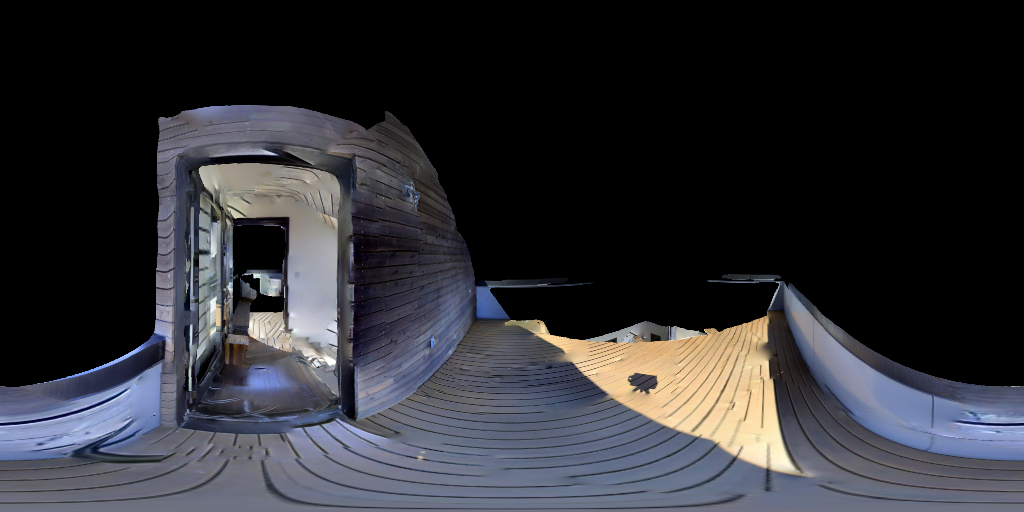} 
    \end{minipage} & \begin{minipage}{0.8\columnwidth}
      \includegraphics[width=\columnwidth]{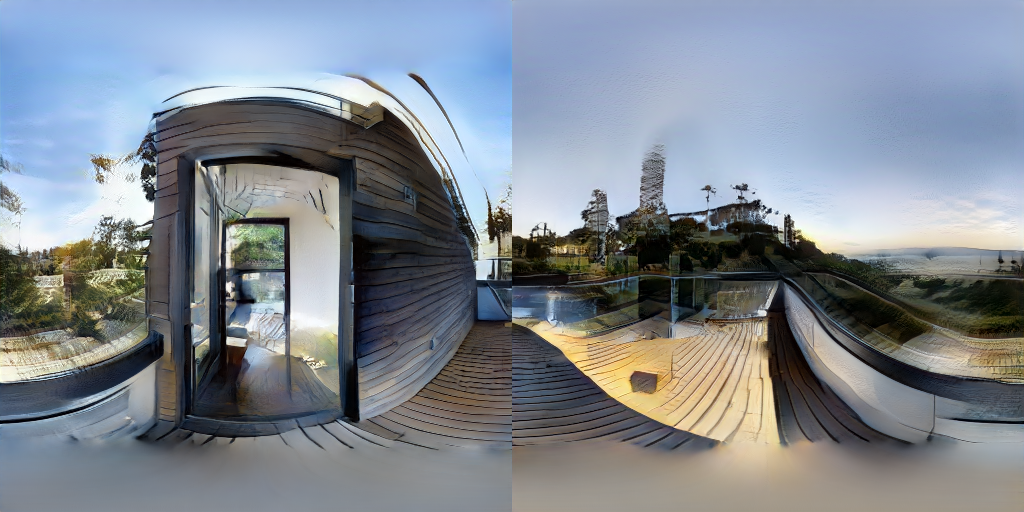} 
    \end{minipage}  \\
    \hline
    5 &   Gibson &  \begin{minipage}{0.8\columnwidth}
   \includegraphics[width=\columnwidth]{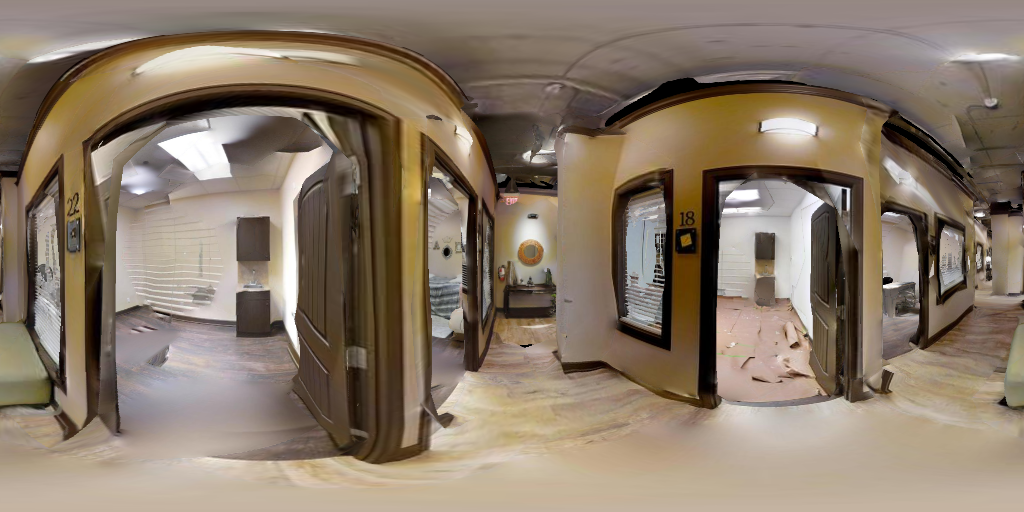} 
    \end{minipage} & \begin{minipage}{0.8\columnwidth}
      \includegraphics[width=\columnwidth]{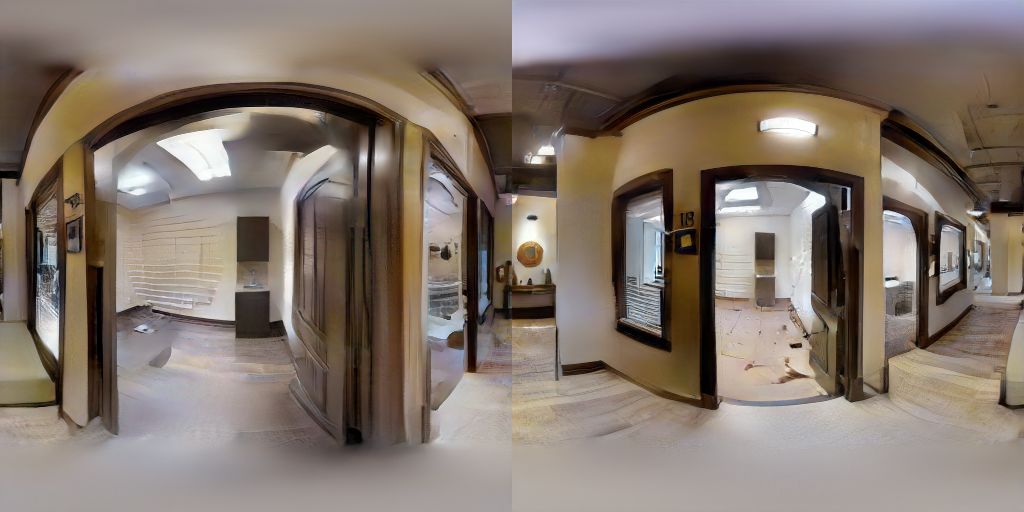} 
    \end{minipage}  \\
    \hline
    6 &   Gibson &  \begin{minipage}{0.8\columnwidth}
   \includegraphics[width=\columnwidth]{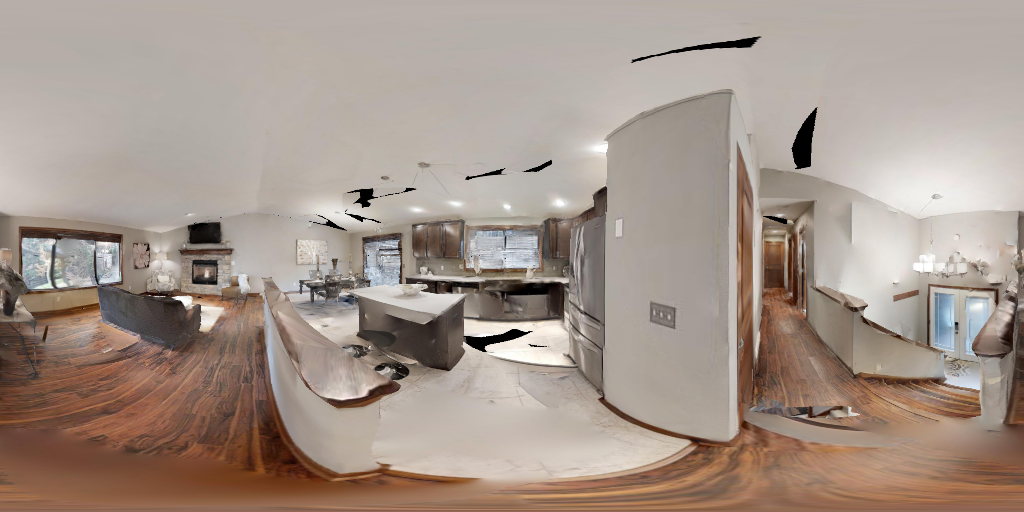} 
    \end{minipage} & \begin{minipage}{0.8\columnwidth}
      \includegraphics[width=\columnwidth]{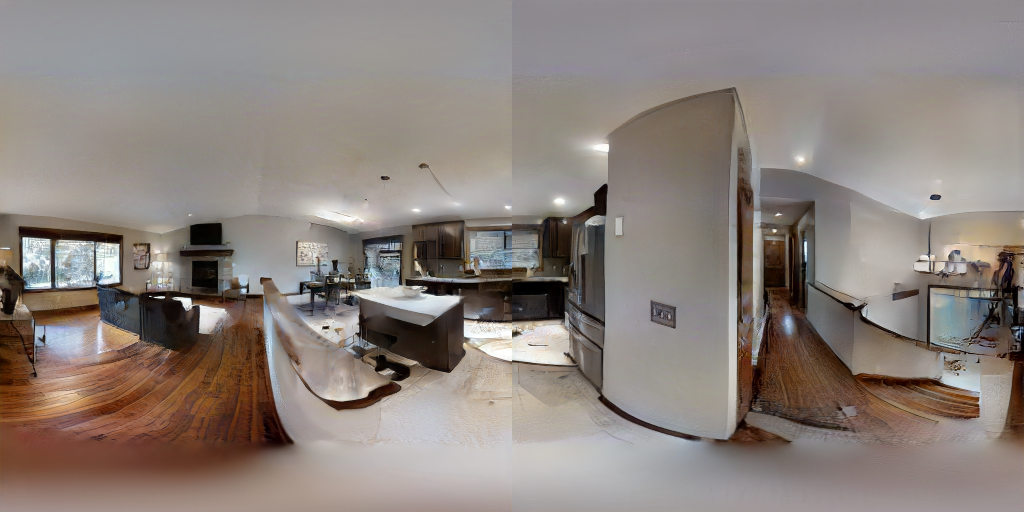} 
    \end{minipage}  \\
    \hline \hline
    \end{tabular}
    }
    \caption{Qualitative examples of our recovered images from HM3D and Gibson environments. The vertical line at the middle of panorama is caused by directly sticking two independently recovered images at 0$^{\circ}$ and 180$^{\circ}$, which will not appear in the resulting augmented data, as explained in Appendix \S\ref{sec_4}.}
    \label{tab:recovered_envs}
\end{table*}

\end{document}